%% file: main.tex
\documentclass[sigconf]{acmart}
\setcopyright{none}

\newcommand{\method}{RGMoE}

\usepackage{enumitem}
\usepackage{wrapfig}
\usepackage{multirow}
\PassOptionsToPackage{table}{xcolor}
\usepackage{xcolor}
\usepackage{hyperref}
\usepackage{url}
\usepackage{enumitem}
\usepackage{multirow}
\usepackage{booktabs}
\usepackage{amsthm}
\theoremstyle{definition}
\newtheorem{problem}{Problem}
\usepackage{adjustbox}
\usepackage{booktabs}
\usepackage{wrapfig}
\usepackage{graphicx}
\usepackage{subcaption} 
\usepackage[table]{xcolor}
\usepackage{booktabs,tabularx,siunitx}
\usepackage{amsmath,amsfonts}
\usepackage{algorithm}
\usepackage{algorithmic}
\begin{document}

\title{Backdoor or Manipulation? Graph Mixture of Experts Can
Defend Against Various Graph Adversarial Attacks}

\author{\textbf{Yuyuan Feng}}
\affiliation{%
  \institution{Xiamen University} 
  \country{}
}
\affiliation{%
  \institution{The Hong Kong University of Science and Technology (Guangzhou)} 
    \country{}
}
\email{30920231154331@stu.xmu.edu.cn}

\author{\textbf{Bin Ma}}
\affiliation{%
  \institution{The Hong Kong University of Science
and Technology (Guangzhou)} %
\country{}
}
\email{binma@hkust-gz.edu.cn}

\author{\textbf{Enyan Dai}}
\authornote{Corresponding author}
\affiliation{%
  \institution{The Hong Kong University of Science
and Technology (Guangzhou)} %
\country{}
}
\email{enyandai@hkust-gz.edu.cn}


\renewcommand{\shortauthors}{Trovato et al.}

\begin{abstract}
Extensive research has highlighted the vulnerability of graph neural networks (GNNs) to adversarial attacks, including manipulation, node injection, and the recently emerging threat of backdoor attacks. However, existing defenses typically focus on a single type of attack, lacking a unified approach to simultaneously defend against multiple threats. In this work, we leverage the flexibility of the Mixture of Experts (MoE) architecture to design a scalable and unified framework for defending against backdoor, edge manipulation, and node injection attacks. Specifically, we propose an MI-based logic diversity loss to encourage individual experts to focus on distinct neighborhood structures in their decision processes, thus ensuring a sufficient subset of experts remains unaffected under perturbations in local structures. Moreover, we introduce a robustness-aware router that identifies perturbation patterns and adaptively routes perturbed nodes to corresponding robust experts. Extensive experiments conducted under various adversarial settings demonstrate that our method consistently achieves superior robustness against multiple graph adversarial attacks. Our code is available at \url{https://github.com/ehappymonkey/RGMoE}.
\end{abstract}



\received{20 February 2007}
\received[revised]{12 March 2009}
\received[accepted]{5 June 2009}

\maketitle

\input{1_intro}
\input{2.5_Prelimaries}

\input{3_Method}

\input{4_Experiments}
\input{2_Related}

\input{5_Conclusion}

\newpage
\bibliographystyle{ACM-Reference-Format}
\bibliography{reference}

\newpage
\appendix

\input{6_Appendix}

\end{document}

%% file: 1_intro.tex
\section{Introduction}


Graphs are ubiquitous data structures in numerous domains, such as social networks~\cite{fan2019graph}, molecular graphs ~\cite{mansimov2019molecular}, and finance systems ~\cite{cheng2022financial}. Recent years, Graph Neural Networks (GNNs)
have shown great success in representation learning on graphs. The success of GNNs relies on the message-passing mechanism. With this mechanism, node representations can capture information of neighbors and graph structure, benefiting various downstream tasks~\cite{fan2019graph, zhou2020graph, scarselli2008graph}.

Although promising results have been achieved, recent studies have shown that GNNs are vulnerable to adversarial attacks. In other words, the predictions of GNNs can greatly degrade or be manipulated under an unnoticeable perturbation in graphs. Based on the way how the graph data is perturbed, the adversarial attacks can be categorized into manipulation attack, node injection attack, and the recent emerging backdoor attack~\cite{dai2024comprehensive}. For instance, PRBCD~\cite{geisler2021robustness} can fool GNNs to give false predictions on target nodes by adding or deleting an unnoticeable amount of edges of the training graph. TDGIA~\cite{zou2021tdgia} manages to significantly reduce the global node classification performance of GNNs by injecting a small amount of labeled fake nodes. Recently, emerging backdoor attacks such as UGBA~\cite{dai2023unnoticeable} and DPGBA~\cite{zhang2024rethinking} can easily control the model’s prediction by injecting specifically designed triggers (i.e., node or a subgraph) into the training graph. During inference, the attacker can make the GNN misbehave when the pre-defined trigger is present.

To address the susceptibility of GNNs to adversarial attacks, many defense methods have been proposed including adversarial training~\cite{xu2019topology}, graph structure denoising~\cite{jin2020graph} and robust aggregation~\cite{chenunderstanding, geisler2021robustness, jin2021node}.
However, these defense strategies have two major limitations: (i) \textit{Incomplete coverage of attacks}: Most existing methods primarily address general manipulation and node injection attacks, but recent studies show that they remain vulnerable to graph backdoor attacks~\cite{dai2023unnoticeable, zhang2024rethinking}. Moreover, the defense against backdoor attacks is rather limited~\cite{zhang2024rethinking, zhang2024robustness}, let alone a unified framework to defend against them simultaneously.  (ii) \textit{Lack of scalability}: Many state-of-the-art defense methods against graph adversarial attacks, such as Pro-GNN~\cite{jin2020graph} and RS-GNN~\cite{dai2022towards}, require dense adjacency matrix computations with a time complexity of $O(N^2)$. Therefore, these methods exhibit poor scalability to large-scale graph datasets.
Given the above limitations, it is essential to design a unified and scalable framework that can effectively defend manipulation attacks, node injection attacks, and backdoor attacks simultaneously.

\begin{figure}[t]
  \centering
  \begin{subfigure}[t]{.23\textwidth}
    \centering
    \includegraphics[width=\linewidth]{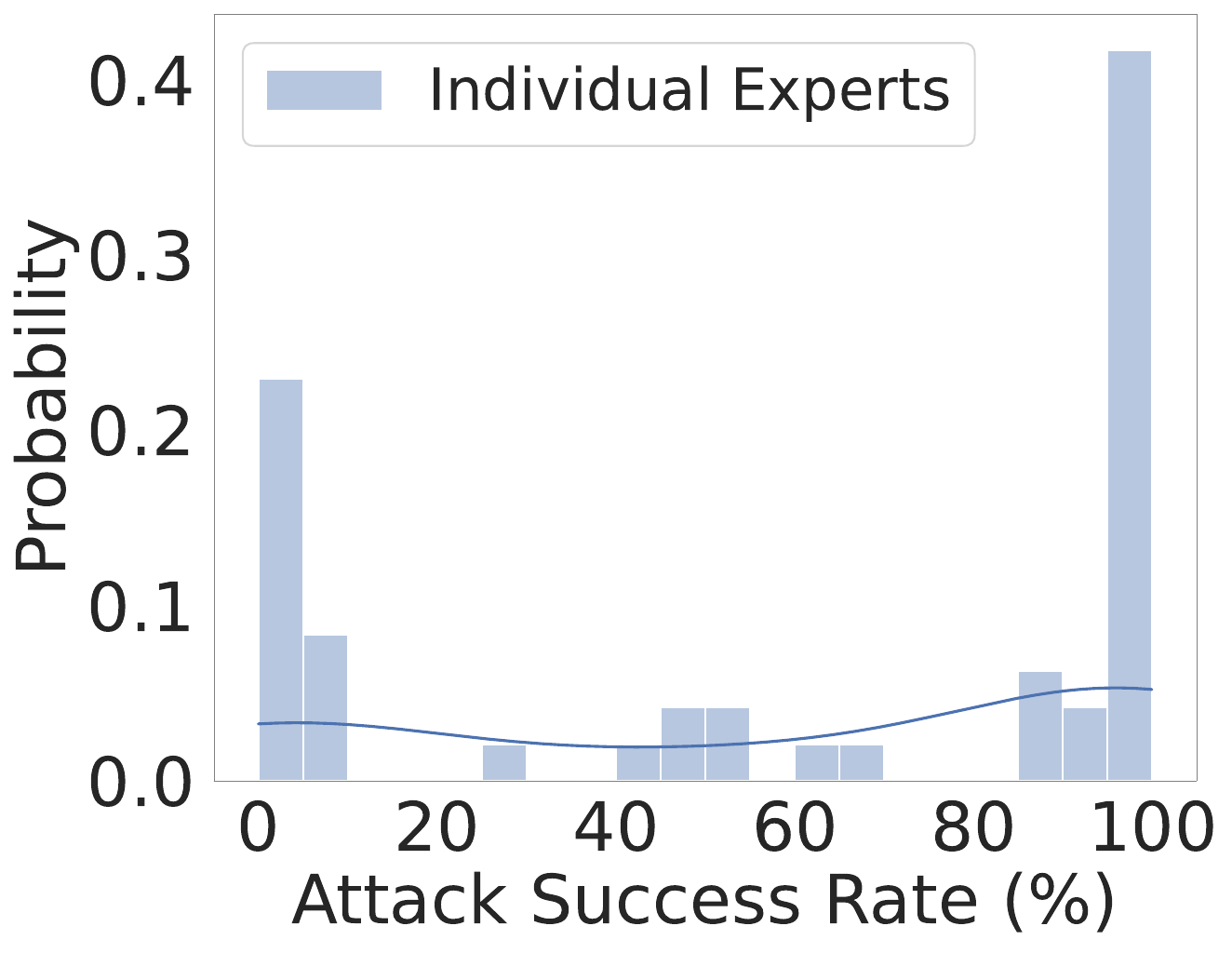}
    \caption{Graph Backdoor Attack}\label{fig:backdoor_experts}
  \end{subfigure}\hfill
  \begin{subfigure}[t]{.23\textwidth}
    \centering
    \includegraphics[width=\linewidth]{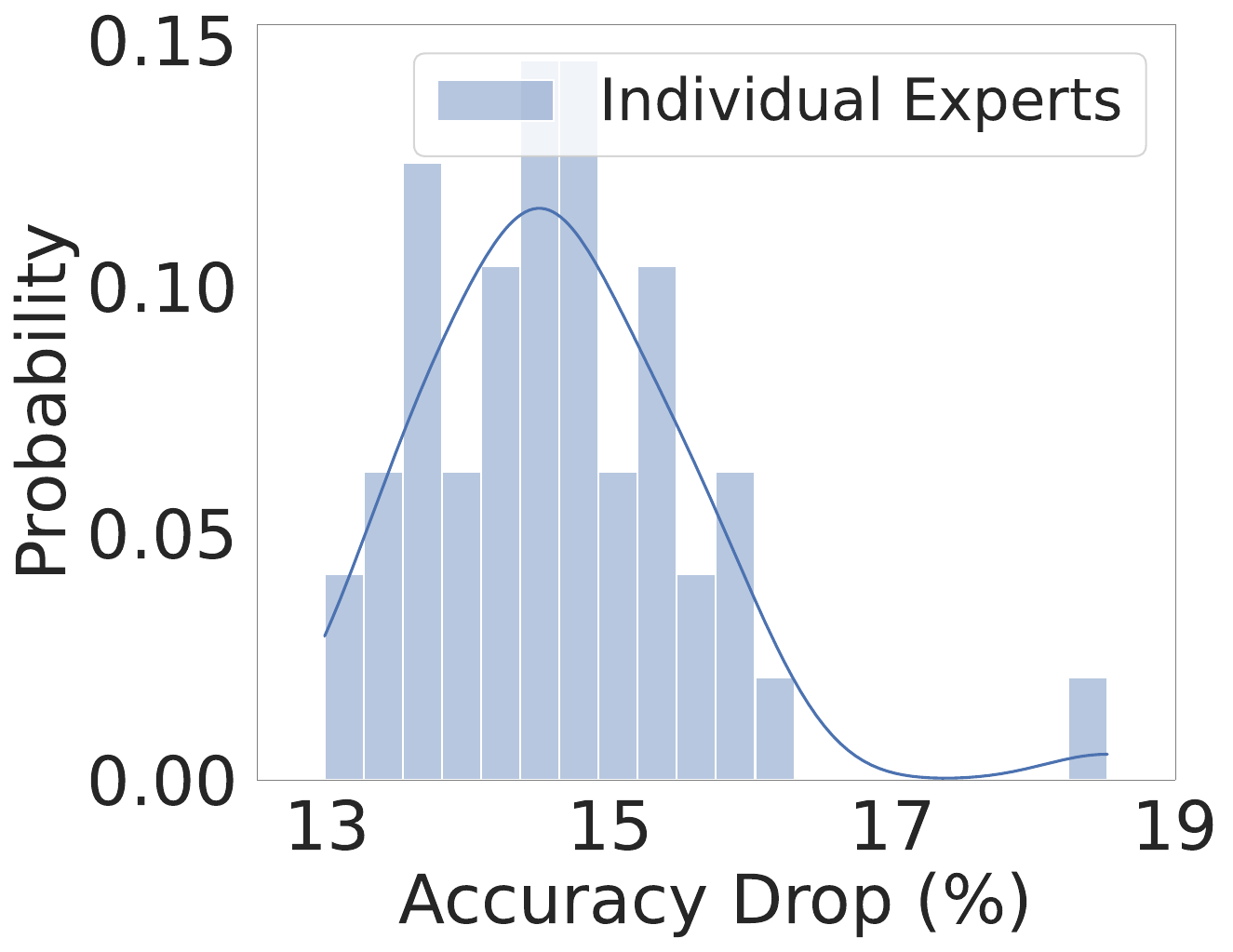}
\caption{Edge Manipulation Attack}\label{fig:prbcd_experts}
  \end{subfigure}\hfill
  \vspace{-0.5em}
  \caption{Performance distribution of individual experts in Graph MoE under various attacks. We report Attack Success Rate (ASR) for backdoor attacks and accuracy drop compared with clean graphs for manipulation attacks.}
  \vspace{-1em}
  \label{fig:distribution}
\end{figure}

One promising direction of defending against various perturbations is to adopt the Mixture of Experts (MoE) \cite{jacobs1991adaptive, wang2023graph, yuan2024mitigating} architecture. \textit{Firstly},  pioneering research has been conducted to promote the diversity of graph experts~\cite{liu2023fair,wu2024graphmetro} for more generalizable graph representations. Inspired by the diversity of experts in MoE, some experts could rely on features unaffected by adversarial attacks, making them naturally robust against the corresponding perturbations. 
As shown in Fig.~\ref{fig:distribution}, in a vanilla graph MoE model, around 20\% experts remain unaffected by backdoor perturbations~\cite{dai2023unnoticeable} on an OGB-Arxiv graph. 
Similarly, we also observe that some experts are more resistant to the PRBCD~\cite{geisler2021robustness} edge manipulation attack. \textit{Secondly}, the flexibility of the routing mechanism allows perturbed nodes to be routed to experts resistant to malicious perturbations, thereby enabling accurate predictions for attacked samples. Moreover, due to sparse routing, the robustness can be achieved without significantly increasing inference-time costs. Therefore, it is promising to learn an efficient Graph MoE with diverse experts and a robustness-aware router to effectively defend against various adversarial attacks.




Two technical challenges remain to be addressed for applying Graph MoE for defending against various adversarial attacks:  
(i) As Fig.~\ref{fig:distribution} shows, the majority of experts still rely heavily on features influenced by adversarial perturbations. Thus, how can we further diversify decision logics of the experts to ensure sufficient number of robust experts for each conducted adversarial attack? (ii) How to learn a robustness-aware router capable of accurately routing perturbed samples to the corresponding experts resistant to the introduced perturbations?
To address these challenges, we propose the \underline{R}obust \underline{G}raph \underline{MoE} ({\method}) to defend various adversarial attacks in a unified framework. Specifically, {\method} incorporates a logic diversity loss to encourage individual experts to focus on different neighbor nodes in their decision process. This diversity would ensure a sufficient number of robust experts for each specific type of potential perturbation. Additionally, {\method} trains a robustness-aware router to adaptively route each perturbed node to the corresponding expert robust to that specific perturbation.
By combining the experts with diverse decision logics and the robustness-aware routing mechanism, our proposed {\method} can effectively defend against various adversarial attacks while keeping the modest inference-time costs.
Our contributions are summarized as follows:

\begin{itemize}[leftmargin=*]
    \item We investigate a novel problem of defending against various adversarial attacks (i.e., backdoor, manipulation, and node injection attack) with a unified framework.
    \item We propose a novel {\method} framework that integrates graph experts with diverse decision logics and a robustness-aware router to defend against various graph adversarial attacks.
    \item Extensive experiments against various adversarial attacks on large-scale graphs demonstrate the effectiveness of our {\method}.
\end{itemize}

%% file: 2.5_Prelimaries.tex
\section{Preliminaries}

Let $\mathcal{G} = (\mathcal{V}, \mathcal{E}, \mathbf{X})$  denote a graph, where $\mathcal{V}$ is the set of $N$ nodes, $\mathcal{E} \subseteq \mathcal{V} \times \mathcal{V}$ is the set of edges, and $\mathbf{X} \in \mathbb{R}^{|\mathcal{V}| \times d}$ represents the set of node features. 
The nodes $\mathcal{V}$ can be divided into: $\mathcal{V}_C$ representing clean nodes, $\mathcal{V}_P$ representing poisoned nodes (i.e., affected by backdoor trigger, manipulated edge, or injected nodes) and unlabeled $\mathcal{V}_U$.
The target label for the poisoned nodes (if modified) is denoted as $y_t$. 
The hidden representation of a node $v$ is $\mathbf{h}_v$ and $\mathcal{N}(v)$ denotes its neighborhood nodes.

\subsection{Preliminaries of Mixture of Experts and Graph Adversarial Attacks}
\label{sec:attacks on gnn}




\textbf{Sparse Mixture of Experts}.
Sparse mixture of experts~\cite{shazeer2017outrageously} decomposes a dense network into multiple experts $\{E_1, E_2, \dots, E_N \}$.
During inference, only a small subset of experts is activated for each input sample. 
A trainable gating network $f_\phi$  determines the subset of experts to the activated based on the input sample. 
Given the input $\textbf{x}$, the gating network outputs a probability distribution over experts $f_\phi(\mathbf{x}) \in \mathbb{R}^N$. The output of MoE is the weighted sum of the top-$K$ expert predictions:
\begin{equation}
    y = \sum_{k\in \mathcal{K}}f_\phi(\mathbf{x})_k E_k(\mathbf{x}).
\end{equation}
where $\mathcal{K}$ denotes indicates the set of activated top-k expert, and $f_\phi(\mathbf{x})_k$ denotes the gating score for the expert $E_k$.


\vspace{0.2em}
\noindent \textbf{Graph Adversarial Attacks}. Extensive studies have shown that GNNs are vulnerable to adversarial attacks. Based on the way how the graph data is perturbed, the adversarial attacks can be categorized into manipulation attack, node injection attack, and the recent emerging backdoor attack~\cite{dai2024comprehensive}:
\begin{itemize}[leftmargin=*]
    \item  \textbf{Graph Backdoor Attack}: In backdoor attacks, the attacker adds backdoor triggers (i.e., nodes or subgraphs) to a set of selected nodes in the original graph and labels them as target class $y_t$. The attacked GNN model trained on the poisoned graph should classify target nodes attached with the triggers as target class and behave normally for clean nodes without triggers attached.
    \item  \textbf{Manipulation Attack}: In a manipulation attack, an attacker manipulates either the graph structure or node features to decrease the accuracy on the global or locally selected test nodes.
    \item  \textbf{Node Injection Attack}: Different from manipulation attacks that modifies the original graph, node injection attack aims to decrease the performance by injecting malicious nodes into the training graph.
\end{itemize}

\subsection{Robustness Analysis of Graph MoE}
\label{sec:preliminary_analysis}



\textbf{Robustness of Individual Experts}. In this subsection, we first conduct preliminary experiments to investigate the robustness of experts in a vanilla graph MoE model. We train a GraphMoE~\cite{wang2023graph} consisting of a gating network and 48 GNN experts and evaluate their robustness against representative graph adversarial attacks, including the \textbf{backdoor attack UGBA}~\cite{dai2023unnoticeable}, the \textbf{manipulation attack PRBCD}~\cite{geisler2021robustness}, and the \textbf{node injection attack TDGIA}~\cite{zou2021tdgia}. 
Fig.~\ref{fig:distribution} shows the performance of individual experts under backdoor attacks and edge manipulation attacks. From Fig.~\ref{fig:backdoor_experts}, we can observe that around 20\% experts achieve ASR less than 20\%. Similarly, a small portion of experts show significantly less performance drop under the manipulation attack in Fig.~\ref{fig:prbcd_experts}. The same trend is also observed for node injection attacks in Appendix~\ref{app:diversity_distribution}. These justify the potential of learning graph mixture of experts against various graph adversarial attacks. 

\begin{figure}[t]
  \centering
  \begin{subfigure}[t]{.219\textwidth}
    \centering
    \includegraphics[width=\linewidth]{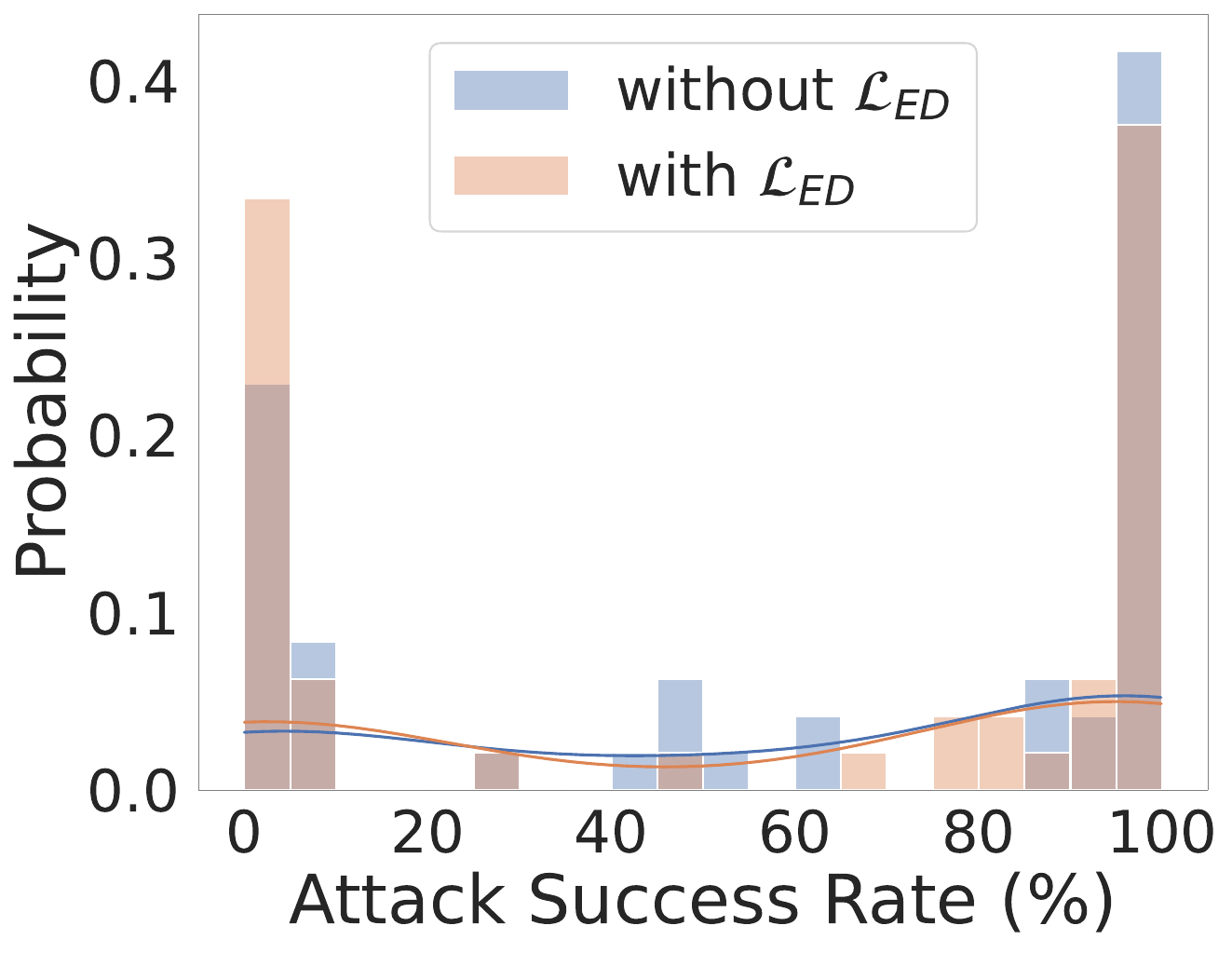}
    \caption{ Robustness Distribution }\label{fig:diverse}
  \end{subfigure}\hfill
  \begin{subfigure}[t]{.22\textwidth}
    \centering
    \includegraphics[width=\linewidth]{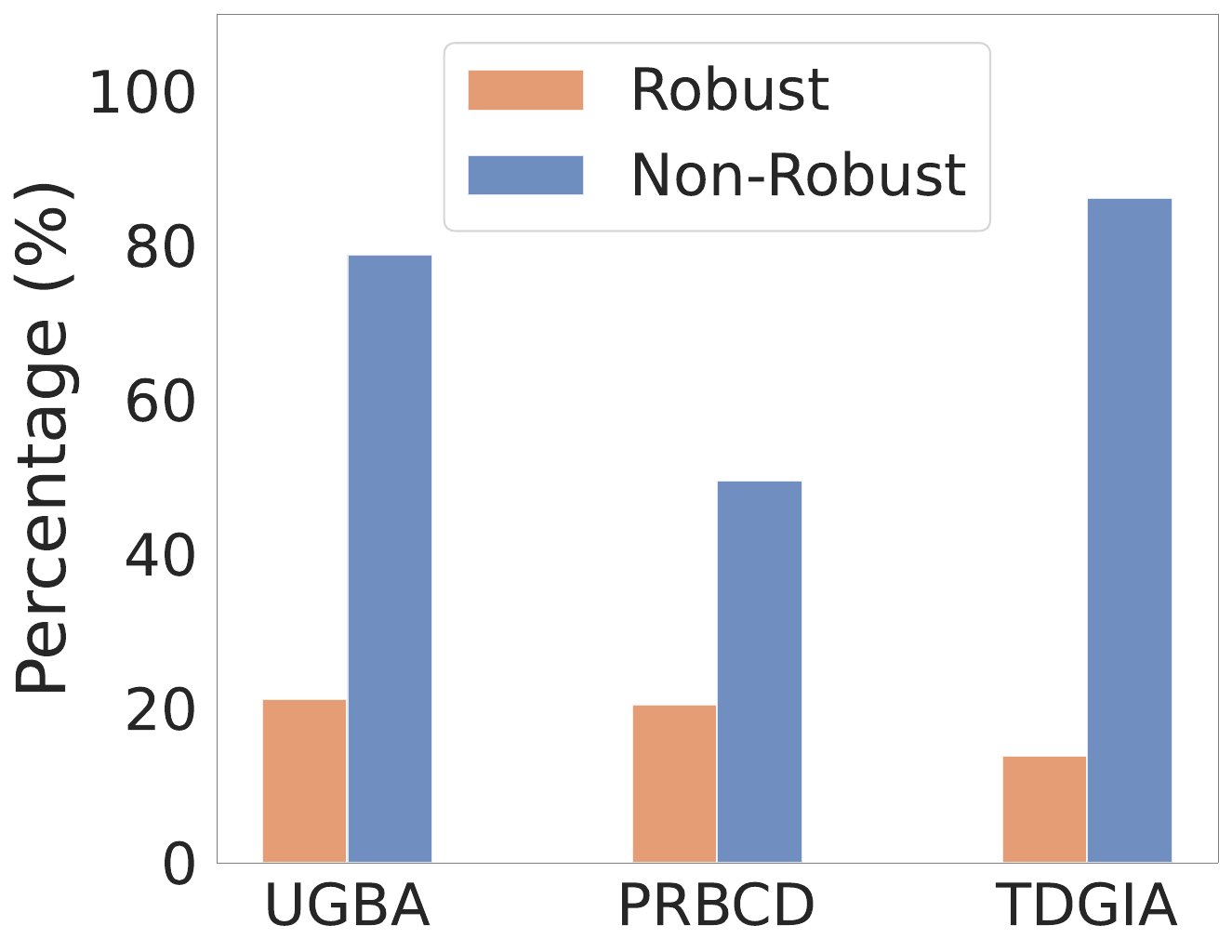}
\caption{Robust Expert Routing Rate}\label{fig:route}
  \end{subfigure}\hfill
  \vspace{-1em}
  \caption{(a) Impacts of existing diversity loss to the proportion of robust individual experts. (b) Rates of routing to roust and non-robust experts. }
  \label{fig:challenges}
  \vspace{-1 em}
\end{figure}

\vspace{0.3em}
\noindent \textbf{Impacts of Diversity Regularization}. 
Although preliminary results show that some experts exhibit robustness, the majority of experts still  rely heavily on features influenced by adversarial perturbations. However, without sufficient number of robust experts, router is very likely to select the non-robust experts. Thus, we explore how existing expert diversity losses affect expert robustness. Specifically, we testify the representation diversification loss $\mathcal{L}_{ED}$ proposed in~\cite{liu2023fair} on OGB-Arxiv under backdoor attacks in Fig.~\ref{fig:diverse}.
 From the Fig.~\ref{fig:diverse}, we can observe a certain degree of improvement with the additional diversify regularization. However, many experts remain vulnerable to adversarial attacks.
This highlights the necessity of exploring more effective approaches to diversify experts' decision logic for robustness to various adversarial attacks.
The detailed formulation of $\mathcal{L}_{ED}$ and results about the robustness distribution under PRBCD and TDGIA can be found in Appendix~\ref{app:diversity_distribution}.

\vspace{0.3em}
\noindent \textbf{Analysis of Routing}. In addition to diverse graph experts robust to different attacks, the Graph MoE requires a router selects the corresponding robust expert for each perturbed node. Otherwise, routing to non-robust experts would still produce wrong predictions. Therefore, we further analyze whether the router trained in Graph MoE can effectively select the corresponding robust experts. Fig.~\ref{fig:route} visualizes the routing rate to robust experts under different attacks. Here, robust experts are defined as those with less than 20\% ASR for backdoor attacks, 14\% ACC drop for manipulation and 5\% ACC drop for node injection attacks. From Fig.~\ref{fig:route}, we observe that although the diversity regularization is beneficial, very few samples (less than 20\%) can be routed to robust experts with the existing router. This highlights the necessity of developing a robustness-aware routing mechanism for the overall robustness of Graph MoE.

\subsection{Problem Formulation}
With the preliminaries of mixture of experts in Sec.~\ref{sec:attacks on gnn} and preliminary analysis in Sec.~\ref{sec:preliminary_analysis}, the problem of learning a robust graph mixture of experts against various adversarial perturbations can be formally defined as:
\begin{problem}
\textit{
Given a graph $\mathcal{G} = (\mathcal{V}, \mathcal{E}, \mathbf{X})$ that may poisoned by various graph adversarial attacks $\mathcal{A}$ (including backdoor attacks, manipulation attacks, and node injection attacks), the robust Graph MoE aims to learn a set of diverse experts 
$\mathcal{M} = \{E_i\}_{i=1}^{K}$, 
where there exists a corresponding subset of experts 
$\mathcal{S}(a(v)) \subseteq \mathcal{M}$ 
that remain unaffected for the sample $v$ perturbed by an attack 
$a \in \mathcal{A}$, 
and a robustness-aware router $f_{\phi}$ 
which can route the perturbed sample $a(v)$ to the corresponding 
robust experts $\mathcal{S}(a(v))$. 
}
\end{problem}


%% file: 3_Method.tex
\section{Methodology}
Our preliminary analysis in Fig.~\ref{fig:distribution} reveals that training a Graph MoE can yield experts identifying different key patterns, with a subset of experts robust to specific adversarial perturbations.
Additionally, the preliminary analysis in Fig.~\ref{fig:challenges} indicates two challenges remain to be addressed: (i) Existing diversity loss is still limited in producing diverse experts robust to different attacks. How can we further diversify decision logics of the experts to ensure the robust experts exist for each conducted adversarial attack? (ii) How can we route perturbed samples to the corresponding robust expert who remains unaffected by the introduced perturbations?  

\begin{figure}[t] 
  \centering
  \includegraphics[width=1.0\linewidth]{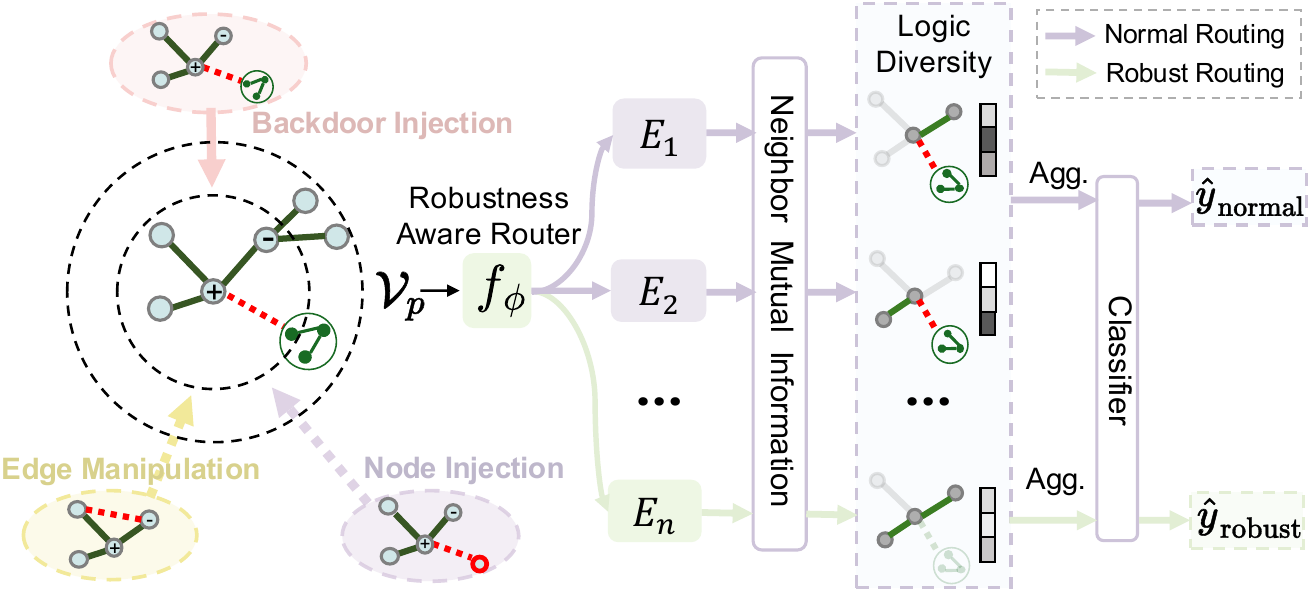}
  \caption{Overall framework of {\method}.}
  \vspace{-1em}
  \label{fig:method}
\end{figure}

To address these challenges, we propose a novel framework {\method}, which is illustrated in Fig.~\ref{fig:method}. As shown in Fig.~\ref{fig:method}, to capture the decision logic of each expert, we measure the contribution of neighbor nodes to the aggregated representation of center node using the mutual information estimator. 
To diversify the prediction logics of experts, {\method} applies a logic diversity loss that encourages different experts to focus on distinct neighbor nodes. This ensures the emergence of a diverse subset of robust experts, each specializing in resistance to specific perturbation patterns. Additionally, {\method} further learns a robustness-aware router to detect the perturbation patterns within samples. Based on these identified patterns, the robustness-aware router adaptively routes each perturbed sample to the corresponding expert that exhibits robustness against that perturbation.
More details of {\method} are given in the following.







\subsection{Mixture of Experts Architecture}
As illustrated in Fig.~\ref{fig:method}, {\method} introduces multiple GNN experts $\{E_1, \dots, E_n\}$ that have distinct decision logics.  The robustness-aware router $f_\phi$ aims to identify the perturbed patterns and activate only the robust experts for each perturbed sample.
In this section, we present the deployed MoE architecture in {\method} that could achieve the above goals.

Following~\cite{wang2023graph}, {\method} deploys a GraphMoE layer comprising multiple graph experts. To determine which experts to use for a given node $v$, a gating function $f_{\phi}$ is employed to sparsely route the top-$k$ experts that are denoted as $\mathcal{K}$. Formally, the representation update within the GraphMoE layer can be written as:
\begin{equation}
    \mathbf{h}(v) = \sum_{k \in \mathcal{K}} f_\phi(v)\text{E}_{k}(v,\mathcal{N}_v),
\label{eq:moe}
\end{equation} 
where each expert $E_{k}$ denotes a graph expert ( e.g., GCN~\cite{kipf2016semi}) using information of neighbor $\mathcal{N}_v$ with message-passing mechanism. 
Since the robustness-aware router require to detect the local structural patterns perturbed by the adversarial attacks, {\method} also uses a GNN as the gating model for router, which also takes into account both the node’s features and neighbors:
\begin{equation}
    f_{\phi}(v) = \text{Softmax}(\text{TopK}(Q(v,\mathcal{N}_v))),
\end{equation}
where $Q$ denotes a separate GNN that can utilize both information of the center node $v$ and its neighbors $\mathcal{N}_v$.

\subsection{Diversifying Expert Decision Logics}
\label{sec:diversify}
In adversarial attacks, attackers perturb neighbors $\mathcal{N}_v$ of center node $v$ through structure manipulation, node injection, or backdoor implantation during the message-passing process, thereby misleading GNN predictions.
To alleviate this vulnerability, we encourage different GNN experts to focus on distinct sets of neighbor nodes, enabling a diverse subset of experts to disregard perturbed patterns.
We first compute the importance of each neighbor node in the message-passing process by measuring the mutual information between neighbor nodes and the aggregated representations of the center nodes. Then, a logic diversity loss is introduced to enforce different experts to rely on distinct local neighborhoods when making predictions. Next, we provide details on how we diversify the logic of each expert.

\subsubsection{Capturing Logic by Mutual Information.}
To capture the decision logic of the expert's message-passing process, we quantify how much each neighbor contributes to the target node’s representation within the message-passing in each expert. Specifically, for each expert $E_k$ in a GraphMoE layer, let $\mathbf{\tilde h}_v^k$ denote the output representation of node $v$, and $\mathbf{h}_{u}^k$ denote the input representation of its neighbor $u \in \mathcal{N}_v$ used by expert $E_k$,
we aim to measure: 
\begin{equation}
    I^k(u,v)=\mathcal{I}(\mathbf{h}_u^{k}; \mathbf{\tilde h}_v^{k}),
\end{equation}
where $\mathcal{I}(\cdot;\cdot)$ denotes mutual information between two variables. 
Since the mutual information quantifies how much information one variable provides about another, $I^k(u,v)$ reflects neighbor $u$'s contribution to the updated representation of node $v$ within expert $E_k$. 
By combining the mutual information contributions of all neighbors, we obtain a decision logic vector that characterizes the message-passing pattern learned by the GNN expert:
\begin{equation}
    \mathbf{I}_{v}^{k} = [I^k(u_1,v), I^k(u_2,v), \dots, I^k(u_{|\mathcal{N}(v)|},v)], \quad u_i \in \mathcal{N}(v).
    \label{eq:decision_logic}
\end{equation}
Since the exact mutual information is intractable, we adopt the Jensen-Shannon divergence-based MI estimator~\cite{nowozin2016f}, denoted by $\hat{\mathcal{I}}_{\omega}^{\text{JSD}}(\mathbf{h}_u^{k};\mathbf{\tilde h}_v^{k})$, to approximate the MI values in Eq.~\ref{eq:decision_logic}:
\begin{equation}
\hat{\mathcal{I}}_\omega^{\text{JSD}}(u;v) := 
\mathbb{E}_{\mathbb{P}_{u,v}}[-\text{sp}(-T_{\omega}(\mathbf{h}_u^{k},\mathbf{\tilde h}_v^{k}))] - 
\mathbb{E}_{\mathbb{P}_u\times\mathbb{P}_v}[\text{sp}(T_{\omega}(\mathbf{h}_u^{k},\mathbf{\tilde h}_v^{k}))],
\label{eq:mutual_information}
\end{equation}
where $\mathbb{P}_{u,v}$ is the joint distribution of neighbor representations and corresponding center-node outputs, and $\mathbb{P}_u\times\mathbb{P}_v$ denotes the product of their marginal distributions. Here, $\text{sp}$ denotes the softplus function.
Following~\cite{nowozin2016f}, We train the discriminator network by maximizing the mutation information:
\begin{equation}
    \min_{\omega} \mathcal{L}_{\text{MI}} = -\frac{1}{|\mathcal{V}|}\sum_{v \in \mathcal{V}}\frac{1}{|\mathcal{N}(v)|}\sum_{u \in \mathcal{N}(v)} \hat{\mathcal{I}}_\omega^{\text{JSD}}(u;v).
    \label{eq:optimize_MI}
\end{equation}



\subsubsection{Logic Diversity Loss.} 
After obtaining each expert's decision logic vector $\mathbf{I}_{v}^{k}$ from Eq.~\ref{eq:decision_logic}, we further diversify these decision logics by minimizing the similarity between pairs of experts. Specifically, we define a logic diversity loss that penalizes high cosine similarity between the decision logic vectors of selected $K$ experts:
\begin{equation}
\mathcal{L}_{\text{logic}} = \frac{2}{|\mathcal{V}|K(K-1)} 
\sum_{v \in \mathcal{V}} \sum_{i=1}^{K-1}\sum_{j=i+1}^{K} \max\left(0,\; s(\mathbf{I}_{v}^{i}, \mathbf{I}_{v}^{j})-m\right),
\label{eq:logic_loss}
\end{equation}
where $s(\cdot, \cdot)$ denotes the cosine similarity between two logic vectors, and $m$ is a margin hyperparameter controlling the degree of diversity among experts. Minimizing this loss encourages each expert to capture distinct local neighborhood structures, thus promoting diversity in experts' decision logics.


\subsubsection{Diverse Experts Training}.
With the above MoE architecture, MI estimator in Eq.~\ref{eq:optimize_MI} and logic diversity loss in Eq.~\ref{eq:logic_loss}, we jointly train the router and experts end-to-end by optimizing:
\begin{equation}
    \min_{\phi, \theta, \omega}\mathcal{L}_{E} =  \mathcal{L}_{MoE} + \mathcal{L}_{MI} + \lambda \mathcal{L}_{\text{logic}},
    \label{eq:phase1}
\end{equation}
where $\mathcal{L}_{\text{MoE}}$ denotes the general MoE loss  denotes the general MoE loss, including the classification loss and the auxiliary load-balancing loss detailed in Appendix~\ref{app:auxiliary_loss}, and $\lambda$ is a hyperparameter controlling the strength of the logic diversity regularization.

\subsection{Robustness-Aware Router Training}
\label{sec:routing} 
With the MI-based logic diversity loss, {\method} can ensure robust experts against adversarial perturbations.  A robustness-aware router is still required to route the poisoned nodes to corresponding robust experts. 
Intuitively, experts tend to exhibit larger disagreement on perturbed nodes, since robust experts rely on features that are unaffected by perturbations, whereas non-robust experts do not. 
This intuition is also empirically supported by Fig.~\ref{fig:distribution}, where robust experts assign significantly different prediction scores to perturbed nodes than non-robust experts. 
Motivated by this, we treat nodes with large expert disagreement as potential perturbed samples. For these potential perturbed nodes, we lower the router's routing confidence toward experts giving consistent predictions, encouraging it to consider experts exhibiting significant disagreement. This enhances the participation of robust experts, whose predictions on perturbed samples typically differ from those of non-robust experts. Next, we present the details of robustness-aware router training. 

\subsubsection{Identification of Potential Perturbed Samples.}
We identify potential perturbed nodes by measuring disagreement among experts in the final GraphMoE prediction layer. Specifically, for a GraphMoE with $N$ experts, let $\mathbf{y}_v^n \in \mathbb{R}^C$ represent the corresponding predictive distribution. The disagreement among experts for node $v$ is quantified as the prediction variance:
\begin{equation}
s(v) = \sum_{n=1}^{N} \text{KL}(\mathbf{y}_v^n \parallel \tilde{\mathbf{y}}_v), \quad 
\tilde{\mathbf{y}}_v = \frac{1}{N}\sum_{k=1}^{N}\mathbf{y}_v^{k},
\label{eq:smooth_label}
\end{equation}
where $\tilde{\mathbf{y}}_v$ is the aggregated prediction probability $y_v^k$ of the expert $E_k$, and $\text{KL}(\cdot \parallel \cdot)$ denotes the Kullback-Leibler divergence.
We empirically observe that perturbed nodes $\mathcal{V}_{P}$ generally exhibit significantly larger expert disagreement than clean nodes $\mathcal{V}_{C}$. Therefore, we define the set of identified potential perturbed nodes $\hat{\mathcal{V}}_P$ as those whose prediction variance exceeds a threshold derived from the distribution of $s(v)$ over all training nodes $\mathcal{V}_L \cup \mathcal{V}_U$.
\begin{equation}
\hat{\mathcal{V}}_P = \{ v \in \mathcal{V}_L \cup \mathcal{V}_{U} \mid s(v) > \mu_s + \sigma_s\},
\label{eq:identify_vp}
\end{equation}
\noindent where $\mu_s$ and $\sigma_s$ denote the mean and standard deviation of $s(i)$ over all nodes. 

\subsubsection{Encouraging to Route to Robust Experts.}  
With the identified perturbed nodes $\hat{\mathcal{V}}_{P}$, we further design an objective to route poisoned nodes to robust experts. 
For these nodes, whose predictions exhibit large disagreement across experts, we lower their confidence in the original prediction and encourage routing toward experts with alternative predictions, i.e., potential robust experts. 
To achieve this, the {\method} adopts a soft label $\tilde{\mathbf{y}}_v$ that aggregates predictions from all experts in the final prediction layer:
\begin{equation}
\tilde{\mathbf{y}}_v = \frac{1}{N}\sum_{k=1}^{N}\mathbf{y}_v^{k},
\end{equation}
This soft label lowers the router's reliance on experts with consistent but potentially non-robust predictions, thus encouraging a diversified routing strategy that leverages robust experts for improved resilience against adversarial perturbations.
Moreover, to further discourage reliance on the original prediction and increase the likelihood of routing to robust experts, we optionally refine the soft label $\tilde{\mathbf{y}}_v$ by applying a shrinkage factor $\rho_c$ (where $0 < \rho_c \leq 1$) to the originally predicted class and then renormalizing:
\begin{equation}
    \tilde{\mathbf{y}}_v(c) = \frac{\rho_c \tilde{\mathbf{y}}_v(c)}{\sum_{j=1}^{C}\rho_j \tilde{\mathbf{y}}_v(j)}, \quad c \in \{1, \dots, C\}.
    \label{eq:smooth_y}
\end{equation}
Here, $\rho_c$ controls the degree of confidence reduction toward the original predicted class, thereby promoting diversified routing to robust experts.
With the above constructed soft label $\tilde{\mathbf{y}}_v$ for perturbed nodes $\hat{\mathcal{V}}_P$ 
and hard label $\mathbf{y}_v$ for clean nodes $\mathcal{V}_C = \mathcal{V}_L \setminus \hat{\mathcal{V}}_P$, 
we train the robustness-aware router by minimizing:
\begin{equation}
\min_{\phi} \mathcal{L}_{\text{router}}
= 
\frac{1}{|\hat{\mathcal{V}}_P|} \sum_{v \in \hat{\mathcal{V}}_P} 
l(\hat{\mathbf{y}}_v, \tilde{\mathbf{y}}_v)
+ 
\gamma \frac{1}{|\mathcal{V}_C|} 
\sum_{v \in \mathcal{V}_C} 
l(\hat{\mathbf{y}}_v, \mathbf{y}_v),
\label{eq:router_obj}
\end{equation}
where $\hat{\mathbf{y}}_v$ denotes the final predictions from the MoE, $l$ denotes the cross-entropy loss, and $\gamma$ controls the relative weight of routing supervision from clean nodes.

\subsection{Final Objective Function}
Combining the diversity loss in Eq.(\ref{eq:logic_loss}) and the training of robustness-aware router by Eq.(\ref{eq:router_obj}), the final objective function 
can be formulated as:
\begin{equation}
\label{eq:final_obj}
\min_{\phi, \theta,\omega} \mathcal{L}_{\text{total}} = \mathcal{L}_{E} + \mathcal{L}_{\text{router}}.
\end{equation}
\noindent The training procedure of {\method} is divided into two phases: we first train the graph MoE model with diverse experts using $\mathcal{L}_{E}$ to obtain robust experts; then we fine-tune the router with $\mathcal{L}_{\text{router}}$ to route the perturbed nodes to these robust experts.
We present the detailed training procedure of {method} in Algorithm~\ref{alg:rgmoe}.

\subsection{Time Complexity Analysis}
In this subsection, we analyze the time complexity of {\method}. Given a graph $\mathcal{G} = \{\mathcal{V}, \mathcal{E}, \mathbf{X} \}$, let $L$ be the number of GCN layers, $h$ the hidden dimension, d the average node degree and $N$ the number of experts. In the training phase, the cost of the router is approximately $O(Lh|\mathcal{V}|d+\mathcal{|V|}N)$ for calculating the message-passing and gating score. The cost of GNN experts is approximately $O(Lh|\mathcal{V}|dK)$, where $K$ is the number of activated experts and $K<N$. According to Eq.~(\ref{eq:mutual_information}), the cost of the MI estimator (i.e., an MLP) is approximately $O(Lh|\mathcal{V}|d)$ by neglecting the number of negative samples, which is a small constant. The cost of calculating the logic diversity loss is approximately $O(L|\mathcal{V}|dK^2)$ according to Eq.~(\ref{eq:logic_loss}). When training the robustness-aware router, the process of calculating expert disagreement is $O(|\mathcal{V}|N)$, which is linear to the number of experts, and the process of constructing the smooth label is $O(|\mathcal{V}|N)$. In the test phase, the cost is only $O(Lh|\mathcal{V}|d+\mathcal{|V|}N)$ for the router and  $O(Lh|\mathcal{V}|dK)$ for the activated experts. Our time complexity analysis proves
that {\method} has great potential in conducting scalable defense on large-scale graphs.

%% file: 4_Experiments.tex
\section{Experiments}

In this section, we evaluate our method on various datasets to answer the following questions:\looseness=-1

\begin{itemize}[leftmargin=*]
\item \textbf{RQ1:} How effective is our method in defending against various GNN adversarial attacks?
\item \textbf{RQ2:} Does scaling the MoE capacity lead to consistent improvements in robustness and accuracy?  
\item \textbf{RQ3:} How do the MI-based diversity loss and robustness-aware router contribute the robustness of {\method}?
\end{itemize}

\begin{table*}[ht]
\centering
\small
\caption{Results under backdoor attacks. We report the attack success rate (ASR) and accuracy on clean samples.}
\vspace{-1em}
\begin{tabular}{ll|cc|cc|cc|cc}
\toprule
\multirow{2}*{\textbf{Attacks}} & \multirow{2}*{\textbf{Defense}} & \multicolumn{2}{c}{\textbf{Cora}} & \multicolumn{2}{c}{\textbf{Pubmed}}  & \multicolumn{2}{c}{\textbf{Flickr}} & \multicolumn{2}{c}{\textbf{OGB-arxiv}} \\
\cmidrule(r){3-4} \cmidrule(r){5-6} \cmidrule(r){7-8} \cmidrule(r){9-10} 
& & ASR (\%) $\downarrow$ & ACC(\%) $\uparrow$  & ASR (\%) $\downarrow$ & ACC (\%) $\uparrow$ & ASR (\%) $\downarrow$ & ACC (\%) $\uparrow$ & ASR (\%) $\downarrow$ & ACC (\%) $\uparrow$ \\
\midrule

\multirow{8}{*}{GTA} 
 & GCN  & $100.00 \pm 0.00$ & $89.14 \pm 0.46$ & $99.76 \pm 0.03$ & $84.29\pm 0.33$ & $100.00 \pm 0.00$ & $38.21 \pm 0.35$ & $95.80 \pm 0.33$ & $57.45 \pm 0.48$ \\
   & GNNGuard & $98.81 \pm 0.37$ & $79.26 \pm 0.60$ & $59.97 \pm 2.32$ & $80.65 \pm 0.13$ & $58.71 \pm 38.58$ & $40.69 \pm 0.25$ & $40.51 \pm  4.80$ & $60.07 \pm 1.76$ \\
& SoftMedian & $23.99 \pm 1.47$ & \underline{89.63 $\pm$ 0.30} & $22.57 \pm 0.20$ & $82.78 \pm 0.12$ & $22.76 \pm 0.02$ & $40.32 \pm 0.00$ & $40.96 \pm 0.66$ & $60.63 \pm 0.68$ \\
 & Prune  & $4.74 \pm 0.21$ & $88.77 \pm 1.06$ & $5.98 \pm 2.08$ & $84.39 \pm 0.09$ & $0.98 \pm 0.02$ & $41.10 \pm 0.19$ & $\underline{8.57 \pm 0.43}$ & $\underline{61.06 \pm 0.46}$ \\
 & OD  & $32.89 \pm 26.51$ & $66.54 \pm 1.67$ & $19.27 \pm 17.15$ & $84.71 \pm 0.32$ & $2.67 \pm 3.77$ & $40.45 \pm 0.12$ & $11.34 \pm 4.77$ & \textbf{61.26 $\pm$ 0.21} \\
 & RIGBD  & $\underline{2.07 \pm 2.93}$ & $87.65 \pm 1.49$ & $2.36 \pm 2.76$ & $83.68 \pm 0.43$ & $\underline{0.00 \pm 0.00}$ & $\underline{41.23 \pm 0.14}$ & $49.75 \pm 8.39$ & $59.35 \pm 0.70$ \\
 & DP-MoE  & $71.47 \pm 21.03$ & $86.54 \pm 3.66$ & $76.89 \pm 15.08$ & $\underline{85.80 \pm 3.84}$  & $70.20 \pm 36.88$ & $34.45 \pm 2.29$ & $30.68 \pm 24.64$ & $50.34 \pm 5.39$ \\
& Ours  & $\textbf{0.74 $\pm$ 0.76}$ & $\textbf{89.68 $\pm$ 3.50}$  & $\textbf{1.52 $\pm$ 0.94}$ & \textbf{87.47 $\pm$ 0.22} & \textbf{0.00 $\pm$ 0.00} & \textbf{42.11 $\pm$ 2.1} & \textbf{3.24 $\pm$ 1.42} & $57.50 \pm 1.36$ \\

\midrule
\multirow{8}{*}{UGBA} 
 & GCN  & $98.67\ \pm 0.00$ & $\underline{89.90 \pm 0.30}$ & $92.88 \pm 1.53$ & $ 80.25 \pm 0.24$  & $99.96 \pm 0.01$ & $39.53 \pm 0.21$ & $99.60 \pm 0.03$ & $\underline{62.47 \pm 0.43}$  \\
   & GNNGuard & $99.41 \pm 0.21$ & $80.74 \pm 0.00$ & $98.67 \pm 1.61$ & $82.02 \pm 0.37$ & $98.89 \pm 0.12$ & $\underline{43.05 \pm 0.26}$ & $98.41 \pm 0.36$ & $58.30 \pm 4.25$ \\
& SoftMedian & $29.63 \pm 2.55$ & $80.37 \pm 0.52$ & $13.88 \pm 0.11$ & $82.65 \pm 0.21$ & $15.95 \pm 1.92$ & $40.32 \pm 0.00$ & $53.06 \pm 1.77$ & $61.49 \pm 0.29$ \\
 & Prune & $97.04 \pm 1.17$ & $87.78 \pm 0.60$  & $93.44 \pm 1.51$ & $82.85 \pm 0.15$ & $99.85 \pm 0.03$ & $40.48 \pm 0.01$  & $99.14 \pm 0.24$ & $61.58 \pm 0.65$ \\
 & OD & $36.00 \pm 34.17 $ & $87.53 \pm 0.30$ & $33.01 \pm 6.70$ & $\underline{83.22 \pm 0.37}$ & $2.29 \pm 3.07$ & $41.17 \pm 0.02$ & $64.12 \pm 36.55$ & $60.49 \pm 0.47$ \\
  & RIGBD & $\underline{5.67 \pm 1.43}$ & $87.90 \pm 1.06$ & $\underline{5.81 \pm 3.71}$ & \textbf{83.83 $\pm$ 0.52} & $\underline{0.00 \pm 0.00}$ & $41.33 \pm 0.11$ & $\underline{0.11 \pm 0.08}$ & $60.68 \pm 0.55$ \\
 & DP-MoE   & $84.74 \pm 4.39$ & $80.99 \pm 1.97$ & $93.63 \pm 2.36$ & $81.21 \pm 2.36$ & $99.02 \pm 0.18$ & $39.42 \pm 1.12$ & $83.74 \pm 2.12$ & $59.14 \pm 6.95$ \\
 & Ours  & \textbf{1.48 $\pm$ 1.51}  & \textbf{89.75 $\pm$ 1.06} & \textbf{1.57 $\pm$ 1.70} & $82.61 \pm 1.70$  & \textbf{0.00 $\pm$ 0.00} & \textbf{45.71 $\pm$ 0.53} & \textbf{0.03 $\pm$ 0.03} & \textbf{63.38 $\pm$ 0.50}  \\
 
\midrule
\multirow{8}{*}{DPGBA} 
 & GCN  & $89.63\pm 2.48$ & $88.02 \pm 0.17$ & $90.65 \pm 1.71$ & $83.26 \pm 0.13$ & $99.98 \pm 0.01$ & $38.60 \pm 0.04$ & $95.92 \pm 0.19$ & $60.78 \pm 1.05$ \\
  & GNNGuard & $37.48 \pm 2.00$ & $79.59 \pm 0.80$ & $49.16 \pm 20.32$ & $\underline{83.49 \pm 0.21}$ & $84.61 \pm 6.81$ & $\underline{43.03 \pm 0.91}$ & $96.39 \pm 0.17$ & $60.26 \pm 1.46$ \\
& SoftMedian & $31.70 \pm 0.55$ & $80.37 \pm 0.30$ & $21.66 \pm 2.08$ & $82.82 \pm 0.15$ & $99.06 \pm 0.09$ & $40.32 \pm 0.00$ & $39.59 \pm 0.51$ & $\underline{61.88 \pm 0.44}$ \\
 & Prune  & $93.77 \pm 1.43$ & $85.49 \pm 0.46$ & $53.84 \pm 0.65$ & $82.04 \pm 0.27$ & $65.01 \pm 45.24$ & $37.36 \pm 4.14$ & $21.31 \pm 0.42$ & $61.49 \pm 0.56$ \\
 & OD    & $94.96 \pm 1.34$ & $81.99 \pm 0.46$ & $88.65 \pm 1.95$ & $83.27 \pm 0.15$ & $10.94 \pm 15.48$ & $40.36 \pm 0.02$ & $98.33 \pm 0.17$ & $58.59 \pm 0.45$ \\
 & RIGBD & $\underline{0.47 \pm 0.33}$ & $86.42 \pm 0.46$ & $\underline{1.81 \pm 0.19}$ & $83.22 \pm 0.48$ & $\underline{0.00 \pm 0.00}$ & $40.96 \pm 0.38$ & $\underline{0.74 \pm 0.79}$ & $61.14 \pm 0.55$ \\
& DP-MoE   & $89.78 \pm 0.63$ & $\underline{88.15 \pm 1.51}$ & $91.59 \pm 0.29$ & $79.91 \pm 0.38$ & $99.26 \pm 0.10$ & $40.77 \pm 0.48$ & $99.75 \pm 0.11$ & $61.60 \pm 0.49$ \\
& Ours  & \textbf{0.30 $\pm$ 0.42} & \textbf{88.27 $\pm$ 4.06} & \textbf{0.62 $\pm$ 0.33} & \textbf{83.75 $\pm$ 0.85} & \textbf{0.00 $\pm$ 0.00} & \textbf{45.14 $\pm$ 2.55} & \textbf{0.02 $\pm$ 0.02} & \textbf{62.11 $\pm$ 0.48} \\
\bottomrule
\end{tabular}
\label{tab:backdoor_results}
\vspace{-0.5em}
\end{table*}

\subsection{Experimental Settings}

\subsubsection{Datasets.}
To validate the effectiveness of our proposed methods, we conduct experiments on four real-world benchmark datasets widely used for node classification, i.e., Cora \emph{(small-scale)}, Pubmed \emph{(medium-scale)} \cite{sen2008collective}, Flickr \emph{(medium-scale)} \cite{zeng2019graphsaint} and OGB-Arxiv \emph{(large-scale)} \cite{hu2020open}. The statistics of the datasets are in Appendix \ref{app:datasets}.

\subsubsection{Attack Methods.} To demonstrate the robustness of our {\method} against graph backdoor, manipulation, and node injection attacks, we employ the following representative attack methods for each type of graph adversarial attack:
\begin{itemize}[leftmargin=*]
    \item \textbf{Graph Backdoor Attack}: we evaluate our method on three state-of-the-art graph backdoor attack methods, i.e., \textbf{GTA}~\cite{zhang2021backdoor}, \textbf{UGBA}~\cite{dai2023unnoticeable}, and \textbf{DPGBA}~\cite{zhang2024robustness}. \textbf{GTA} is the first backdoor attack on GNNs that defines the graph-oriented triggers as specific subgraphs. The backdoored GNN should predict the target label for any test samples attached with the trigger and behave normally otherwise. \textbf{UGBA} selects representative nodes as poisoned nodes and optimizes the adaptive trigger generator with similarity-constrained loss to conduct unnoticeable graph backdoor attacks with a limited attack budget. \textbf{DPGBA} further introduces adversarial learning to generate in-distribution triggers.
    \item \textbf{Manipulation Attack}: We evaluate \textbf{PRBCD}~\cite{geisler2021robustness}. PR-BCD fools the GNN to give false predictions on target nodes by adding or deleting an unnoticeable number of edges. Its attack maintains efficiency without scaling quadratically in the number of nodes. It has been proven to deliver scalable and powerful attacks.
    \item \textbf{Node Injection Attack}: We examine \textbf{TDGIA}~\cite{chenunderstanding}, a powerful graph injection attack that injects a small amount of labeled fake nodes progressively around
topologically vulnerable nodes in the graph. The authors
introduced homophily unnoticeability to promote unnoticeability while maintaining the damage.
\end{itemize}

\subsubsection{Baseline Methods.}
To evaluate the performance of our method, we compare against three categories of baseline approaches.

\noindent \textbf{Backdoor Defense Methods.} We compare with recent backdoor defense strategies: \textbf{Prune}~\cite{dai2023unnoticeable} detects and removes triggers by identifying edges with low cosine similarity; \textbf{OD}~\cite{zhang2024rethinking} uses a graph auto-encoder to filter out-of-distribution trigger nodes based on reconstruction loss; \textbf{RIGBD}~\cite{zhang2024robustness} leverages random edge dropping to detect backdoored samples with larger prediction variance, achieving state-of-the-art performance.

\noindent \textbf{Robust GNN Methods.} We also compare with robust GNN approaches designed to defend against adversarial attacks: \textbf{RGCN}~\cite{zhu2019robust} employs Gaussian distributions with variance-based attention to absorb adversarial effects; \textbf{GNNGuard}~\cite{zhang2020gnnguard} deletes adversarial edges based on node similarity and adjusts edge weights during training; \textbf{SoftMedian}~\cite{geisler2021robustness} uses weighted mean aggregation based on distance to dimension-wise median; \textbf{SimPGCN}~\cite{jin2021node} adaptively integrates graph structure and node features through similarity-preserving aggregation.

\noindent \textbf{MoE-based Defense.} Finally, we compare with \textbf{DP-MoE}~\cite{yuan2024mitigating}, which combines differential privacy and MoE to apply adaptive noises for counteracting features affected by injected nodes.

\vspace{-0.5em}

\subsubsection{Evaluation Metrics.} Here, we present the details of evaluating the robustness against various graph adversarial attacks.

\noindent \textbf{Graph Backdoor Attacks}.
Following the existing representative graph backdoor studies \cite{dai2023unnoticeable, zhang2024robustness}, we conduct experiments on the inductive node classification task, where 20\% of the nodes are masked out from the original graph as test nodes. 
The attacker can inject 5\% nodes out of the original graph as poisoned nodes on Cora, Pubmed, Flickr, and 3\% on Arxiv. The target class is set to 0, and the trigger size is limited to 3 on all datasets. During defense, we attach half of the test nodes (10\%) with backdoor triggers to test the \textbf{attack success rate (ASR)}. The remaining 10\% are kept clean to test the clean \textbf{accuracy (ACC)}.

\noindent \textbf{Manipulation and Node Injection Attacks}.
Since PRBCD and TDGIA both aim to decrease the overall performance, we follow the previous study~\cite{chenunderstanding2022} to report the \textbf{accuracy (ACC)} on 20\% test nodes of poisoned graphs. For PRBCD, we manipulate different ratios of edges to evaluate the consistent efficacy of {\method}. We deploy a 2-layer GCN as the model architecture across all experiments. Each experiment is conducted 3 times and the average results are reported. More implementation details and the attack budget for TDGIA are specified in Appendix \ref{app:implement_details}.

\begin{table*}[t]
\small
\centering
\caption{Comparison with baselines in defending against manipulation attacks.}
\vspace{-1em}
\begin{tabularx}{0.98\linewidth}{ll *{8}{>{\centering\arraybackslash}X}}
\toprule
\textbf{Ptb} & \textbf{Dataset} & \textbf{GCN (clean)} & \textbf{GCN} & \textbf{RGCN} & \textbf{GNNGuard} & \textbf{SoftMedian} & \textbf{SimPGCN} & \textbf{DP-MoE} & \textbf{Ours}\\
\midrule
\multirow{2}{*}{3\%} & Flickr & 44.21 $\pm$ 0.79 & 42.27 $\pm$ 0.17 & \underline{47.32 $\pm$ 0.24} & $43.96 \pm 1.61$  & 43.31 $\pm$ 0.02 & 42.80 $\pm$ 0.13 & $43.96 \pm 0.02$  & \textbf{51.04 $\pm$ 0.21} \\
 & Arxiv   & 64.45 $\pm$ 0.36 & 48.79 $\pm$ 0.63 & \underline{50.71 $\pm$ 0.15} & $49.06 \pm 0.61$ & 50.46 $\pm$ 0.36 & 50.17 $\pm$ 0.22 & $49.49 \pm 0.65$  & \textbf{53.55 $\pm$ 0.02} \\
\midrule
\multirow{2}{*}{5\%}  & Flickr & -- & 41.32 $\pm$ 0.74 & \underline{46.84 $\pm$ 0.36} & $42.44 \pm 1.74$ & 42.71 $\pm$ 0.01 & 42.80 $\pm$ 0.14  & $43.71 \pm 0.01$ & \textbf{50.98 $\pm$ 0.18} \\
 & Arxiv  & --  & 46.13 $\pm$ 0.55 & 48.10 $\pm$ 0.12 & $46.43 \pm 0.37$ & \underline{48.10 $\pm$ 0.11} & 47.38 $\pm$ 0.21 & $47.27 \pm 0.69$  & \textbf{50.71 $\pm$ 0.09} \\
 \midrule
\multirow{2}{*}{10\%} & Flickr & -- & 38.02 $\pm$ 2.53 & \underline{43.46 $\pm$ 0.77} & $40.43 \pm 2.45 $ & 42.71 $\pm$ 0.00 & 42.79 $\pm$ 0.12  & $43.23 \pm 0.03$ & \textbf{50.16 $\pm$ 0.63}\\
 & Arxiv  & -- & 34.52 $\pm$ 0.44 & 36.59 $\pm$ 0.16 & $34.50 \pm 0.60$ & \underline{37.64 $\pm$ 0.20} & 35.34 $\pm$ 0.21 & $35.72 \pm 0.60$  & \textbf{39.94 $\pm$ 0.04} \\

\bottomrule
\end{tabularx}
\label{tab:prbcd}
\end{table*}

\begin{table*}[t]
\small
\centering
\caption{Comparison with baselines in defending against node injection attacks.}
\vspace{-1em}
\begin{tabularx}{0.98\linewidth}{ll *{7}{>{\centering\arraybackslash}X}}
\toprule
 \textbf{Dataset} & \textbf{GCN (clean)} & \textbf{GCN} & \textbf{RGCN} & \textbf{GNNGuard} & \textbf{SoftMedian} & \textbf{SimPGCN} & \textbf{DP-MoE} &  \textbf{Ours} \\
\midrule
Flickr & 44.21 $\pm$ 0.79 & 43.20 $\pm$ 0.06 & \underline{49.16 $\pm$ 0.07} & $44.91 \pm 1.52$ & 42.71 $\pm$ 0.00 & 42.82 $\pm$ 0.13  & $42.67 \pm 0.04$ & \textbf{49.62 $\pm$ 0.24} \\
Arxiv  & 62.29 $\pm$ 0.39 & 58.79 $\pm$ 0.19 & 61.16 $\pm$ 0.20 & $57.11 \pm 0.61$ & 55.53 $\pm$ 0.34 & 60.53 $\pm$ 4.36  & \underline{$61.76 \pm 0.37$} & \textbf{63.33 $\pm$ 0.46} \\
\bottomrule
\end{tabularx}
\label{tab:tdgia}
\end{table*}

\vspace{-0.7em}
\subsection{Robustness of {\method}}

To answer \textbf{RQ1}, we compare {\method} with baseline defense methods across the datasets under various graph adversarial attacks. 

\noindent \textbf{Defend Against Graph Backdoor Attacks}. 
To investigate the robustness of {\method} to various backdoor attacks, we evaluate its performance across the datasets under three state-of-the-art attack methods, i.e., GTA, UGBA and DPGBA
Following the above-mentioned evaluation protocol, we report the ASR and ACC in 
Table \ref{tab:backdoor_results}.
From Table~\ref{tab:backdoor_results}, we make two key observations: (\textbf{i}) 
Across all baselines, few of them can give effective defenses across all attacks. Another MoE-based architecture, DP-MoE gives poor performance on backdoor attacks with ASR$>$80\% on many datasets.
By contrast, our {\method} consistently achieves the lowest ASR, often approaching 0\% across the datasets and backdoor attack methods. This demonstrates the effectiveness of our {\method} in defending against backdoor attacks. (\textbf{ii}) Our method often attains slightly higher clean accuracy compared to vanilla GCN and most defense baselines. This is because that logic diversity loss encourages different experts to learn complementary local structures. As a result, the ensemble of the diverse experts in Graph MoE facilitate both robustness and accuracy on clean nodes.

\vspace{0.2em}
\noindent \textbf{Defend Against Manipulation and Node Injection Attacks}.
We then evaluate {\method} on large-scale datasets under PRBCD \emph{(edge manipulation)} and TDGIA \emph{(node injection)}. These two attack methods are selected due to their scalability to large-scale datasets. The results of defending against manipulation and node injection attacks on large-scale datasets are presented in Table~\ref{tab:prbcd} and Table~\ref{tab:tdgia}, respectively.  Additional results on small-scale datasets are presented
in Appendix~\ref{app:results_adv}. 
As both PRBCD and TDGIA is designed to degrade the overall performance, we report the ACC (\%) of the test nodes to measure the robustness of {\method} against manipulation and node injection attacks. 
From the tables, we observe: (\textbf{i}) In Table~\ref{tab:prbcd}, {\method} consistently achieves higher accuracy than all baselines across perturbation rates from 3\% to 10\%,  (\textbf{ii}) Similarly in Table~\ref{tab:tdgia}, our method not only surpasses all baselines but in some cases even exceeds the accuracy on the clean graph. These results demonstrate that {\method} provides strong defense against both structural manipulations and node injections.

\vspace{-0.6em}

\subsection{Impact of MoE Capacity to Robustness}

To answer \textbf{RQ2}, we scale the capacity of {\method} by varying the number of experts $N \in \{12, 18, 24, 30, 36, 42, 48\}$, while consistently routing the top $N/6$ experts. This setup allows us to investigate whether increasing MoE capacity leads to improvements in robustness or accuracy. For fair comparison, other parameters (i.e., $\lambda_{div}$ and $\gamma$) are tuned till best. Results across datasets are summarized in Figure~\ref{fig:capacity_ugba}.  More results under other attacks can be found in Appendix 
\ref{app:capacity_results}. 
From the figure, we make two key observations:  
\begin{itemize}[leftmargin=*]
\item Scaling the model capacity greatly enhances robustness. For example, with only $N=12$ experts, the ASR under DPGBA exceeds 25\% on Cora, Pubmed, and Arxiv, with large variance across random seeds. As the number of experts increases, the ASR on all three datasets converges to nearly 0 when $N=48$. This indicates that a larger pool of experts can result in more experts unaffected by backdoor triggers. Consequently, the robustness-aware router can reliably route poisoned nodes to these unaffected experts across different runs. 
\item Dataset scale effect at fixed capacity. At the \emph{same} number of experts, Arxiv exhibits lower ASR than the smaller datasets (Cora and Pubmed). This is because, on larger datasets, MoE can exploit richer data to learn more diverse features that are less aligned with adversarial patterns, yielding stronger robustness. 
\item In contrast, the clean accuracy remains stable or slightly improves with scaling, staying comparable to vanilla GCN. This implies the diversity of experts wouldn't harm the overall accuracy of the MoE framework.
\end{itemize}

\begin{figure}[t]
  \centering
  \begin{subfigure}[h]{.48\columnwidth}
    \centering
    \includegraphics[width=\linewidth]{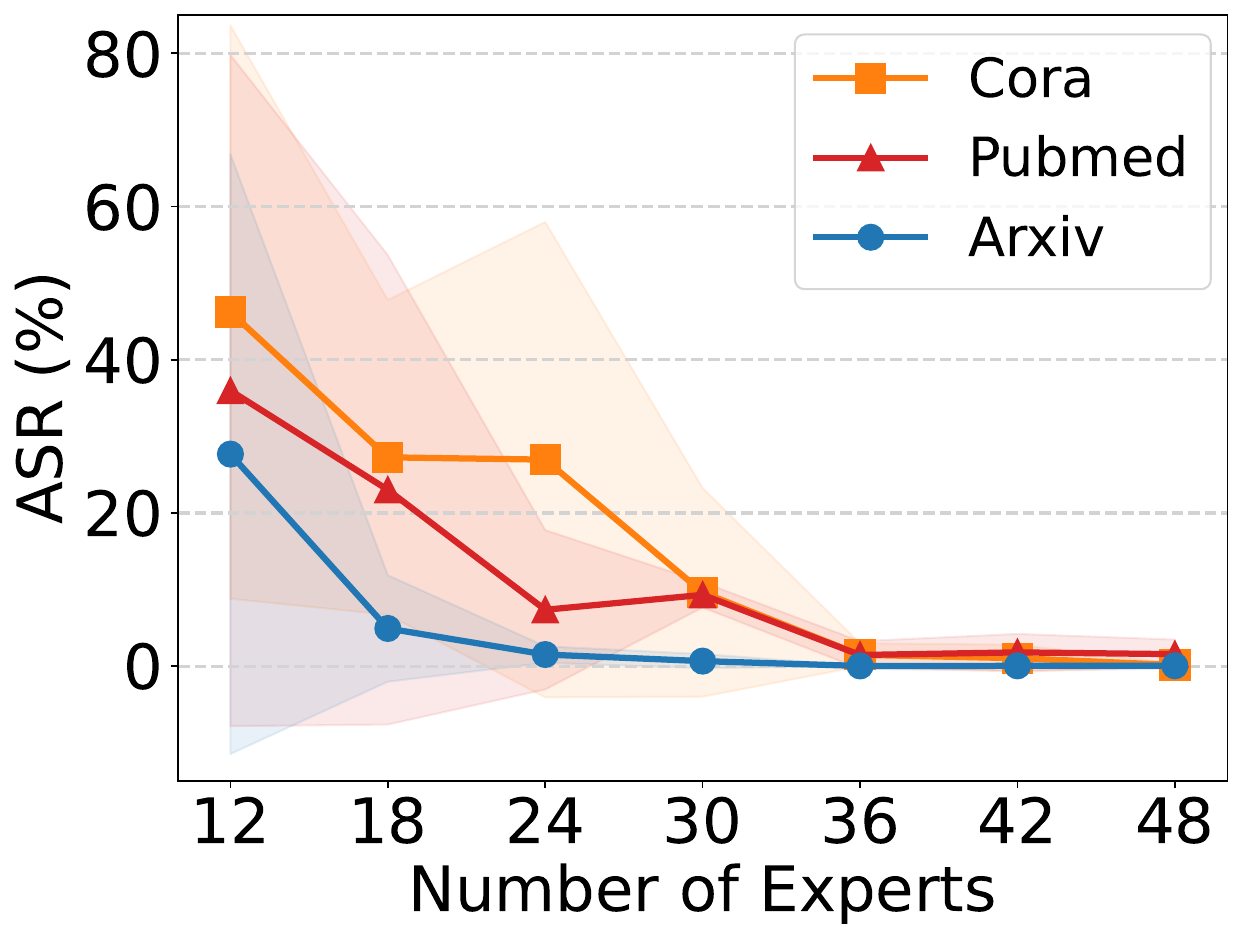}
  \end{subfigure}
  \hfill
  \begin{subfigure}[h]{.48\columnwidth}
    \centering
    \includegraphics[width=\linewidth]{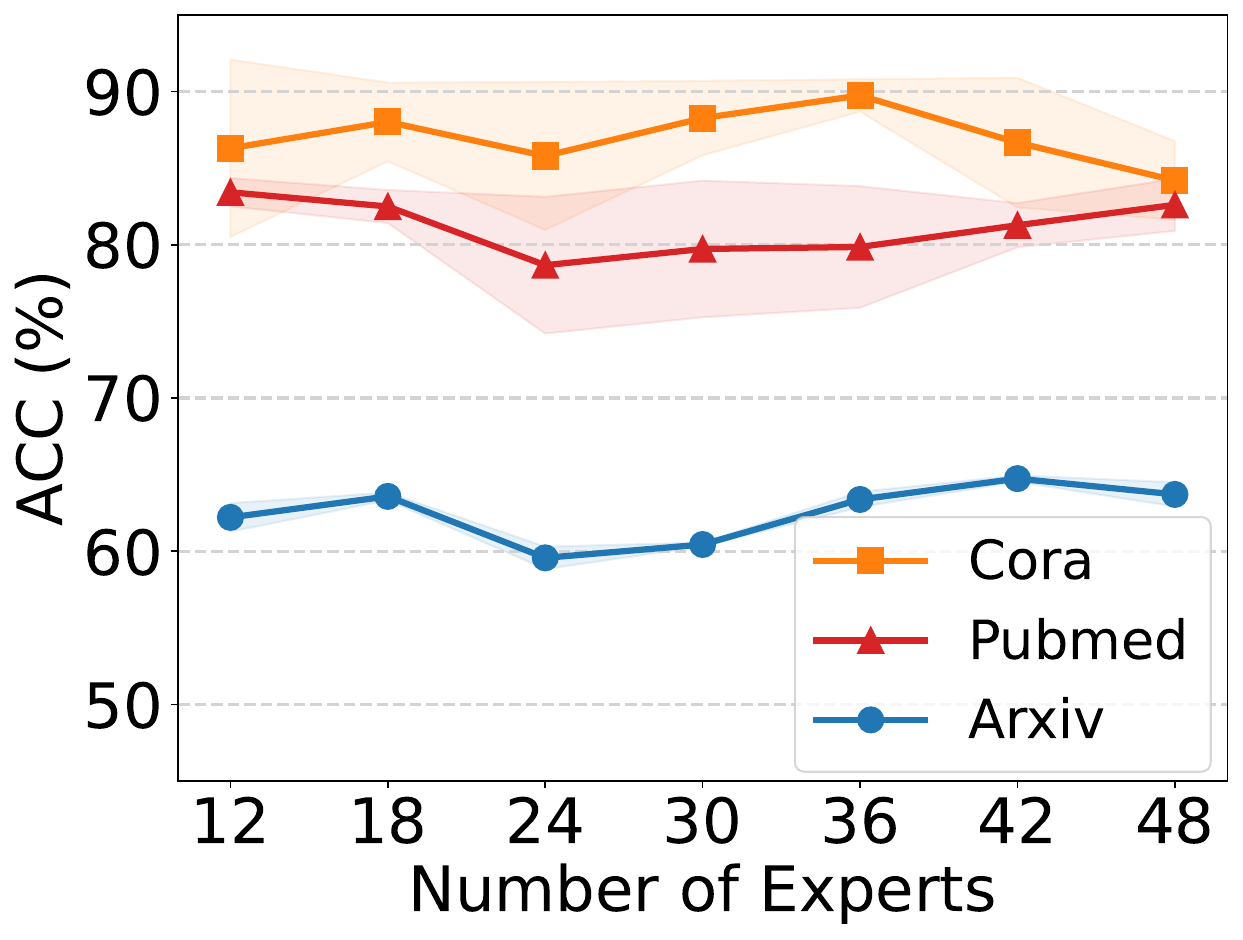}
  \end{subfigure}
  \vspace{-2mm}
  \caption{Impact of model capacity under UGBA attack.}
  \vspace{-1em}
  \label{fig:capacity_ugba}
\end{figure}


\subsection{Ablation Study}

\begin{figure}[t]
  \centering
  \begin{subfigure}[t]{.48\columnwidth}
    \centering
    \includegraphics[width=\linewidth]{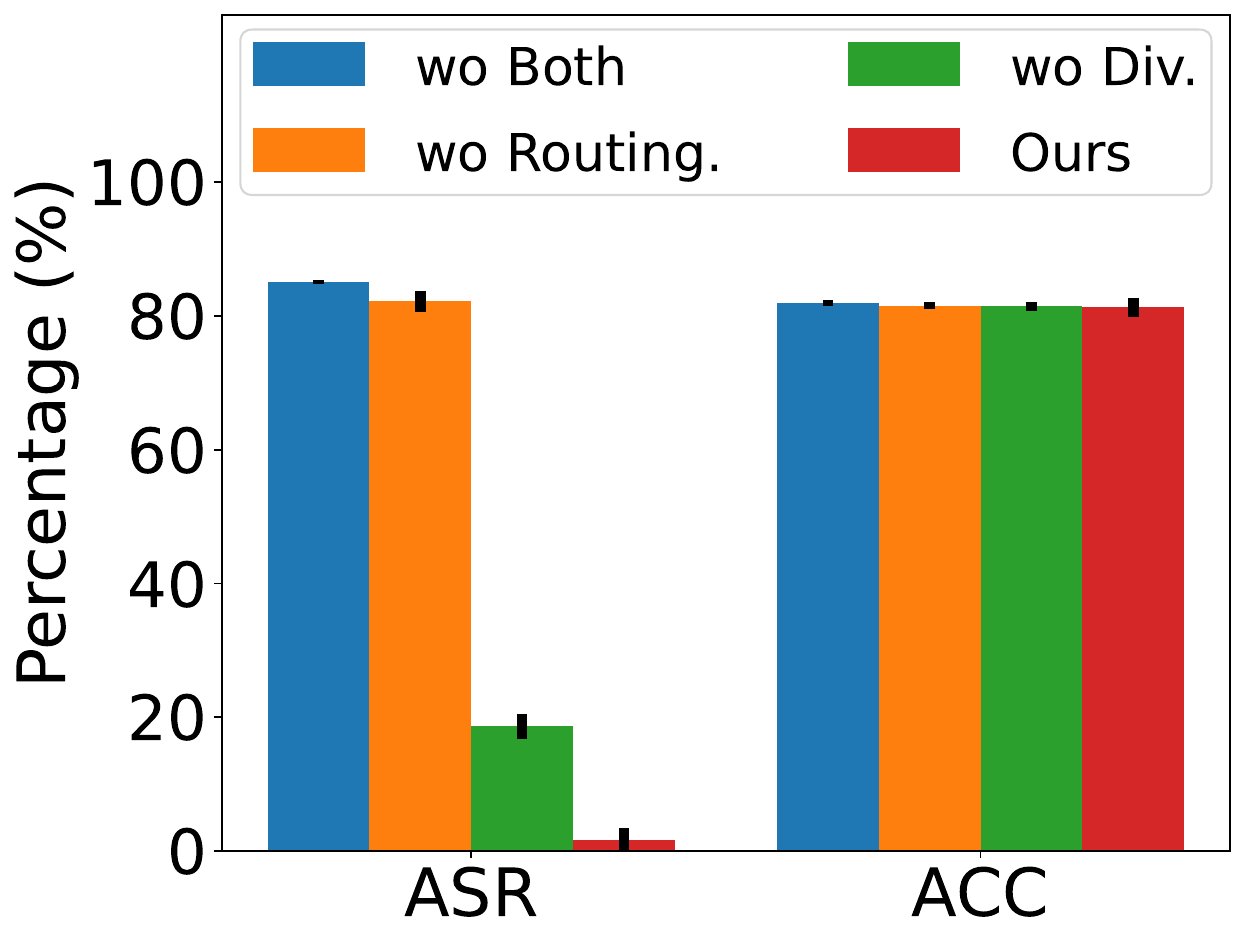}
    \vspace{-1em}
    \caption{Pubmed}\label{fig:abl_cora}
  \end{subfigure}
  \hfill
  \begin{subfigure}[t]{.48\columnwidth}
    \centering
    \includegraphics[width=\linewidth]{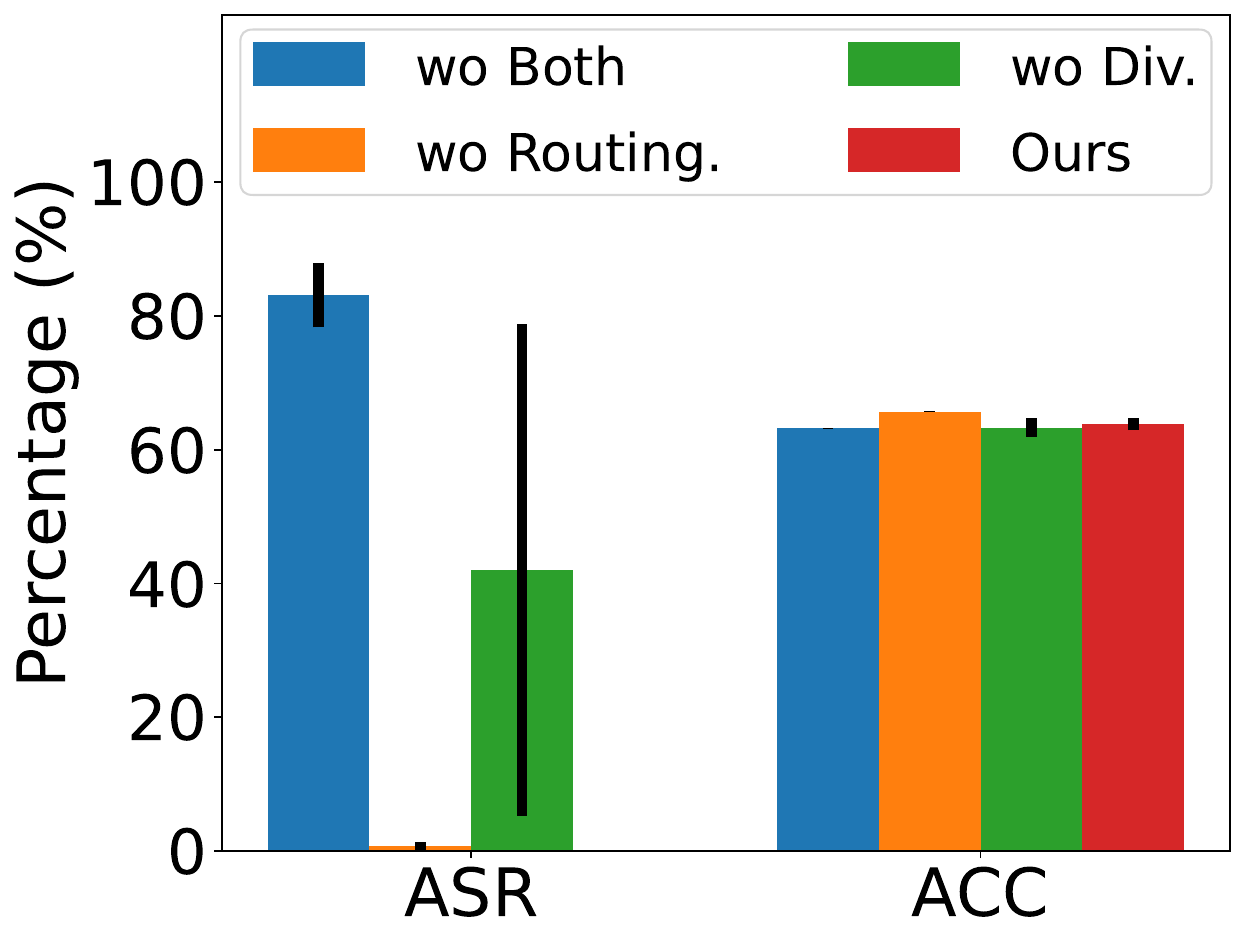}
    \vspace{-1em}
    \caption{Arxiv}\label{fig:abl_arxiv}
  \end{subfigure}
  
  \vspace{-1em}
  \caption{Ablation studies under UGBA across datasets.}
  \label{fig:abl_ugba}
  \vspace{-1em}
\end{figure}

To answer \textbf{RQ3}, we conduct ablation studies to examine the contributions of the proposed logic diversity loss and the robustness-aware routing strategy. Specifically, we (i) set $\lambda_{\text{div}}=0$ to remove the diversity loss, and (ii) replace the robustness-aware routing with random top-$k$ routing, and (iii) removing both the diversity loss and robustness-aware router. Figure~\ref{fig:abl_ugba} reports results on Pubmed and Arxiv under the UGBA attack. Please refer to Appendix \ref{app:ablation_results} for complete datasets and attack methods.
From the figure, we make two observations:  
\begin{itemize}[leftmargin=*]
    \item With $\lambda_{\text{div}}=0$, the ASR drops considerably (e.g., from above 80\% to about 20\% on Pubmed and 40\% on Arxiv), showing that the logic diversity loss is essential in ensuring a number of experts are robust to various attacks, thus enabling the router to select robust experts without scarifying the accuracy.
    \item When removing the robustness-aware router, the attack success rate (ASR) increases significantly on small-scale datasets such as Cora and Pubmed, indicating that the router effectively redirects perturbed nodes to robust experts and thus enhances overall robustness. On large-scale datasets (Flickr and Arxiv), the ASR also shows a relatively smaller increase, since these datasets inherently contain more diverse features that are less aligned with adversarial patterns, resulting in more robust experts that the router can easily utilize.
    \item When combining the robustness-aware router and logic diversity loss, ASR further decreases to nearly 0\% on both datasets. This highlights that the diversity loss and the robustness-aware router are complementary: the loss encourages experts to learn distinct, less-correlated features, while the router ensures that backdoored nodes are directed to the most robust experts. 
\end{itemize}

\begin{figure}[ht]
  \centering
  \begin{subfigure}[t]{.48\columnwidth}
    \centering
    \includegraphics[width=\linewidth]{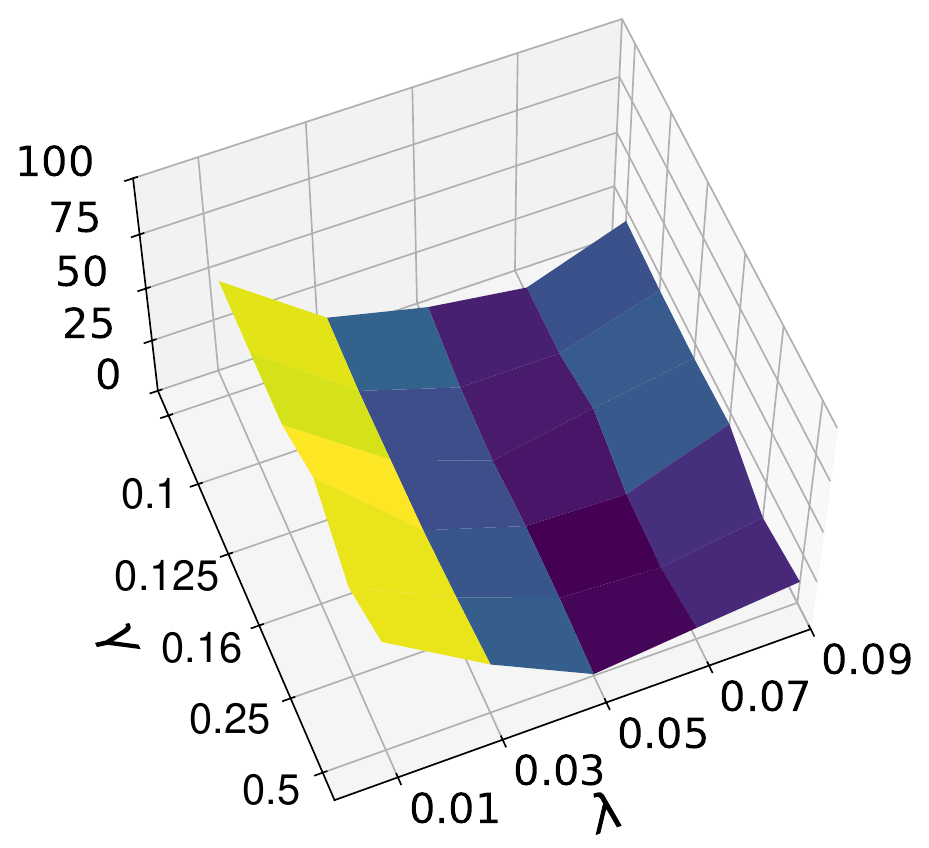}
    \caption{ASR (\%)}
  \end{subfigure}
  \hfill
  \begin{subfigure}[t]{.48\columnwidth}
    \centering
    \includegraphics[width=\linewidth]{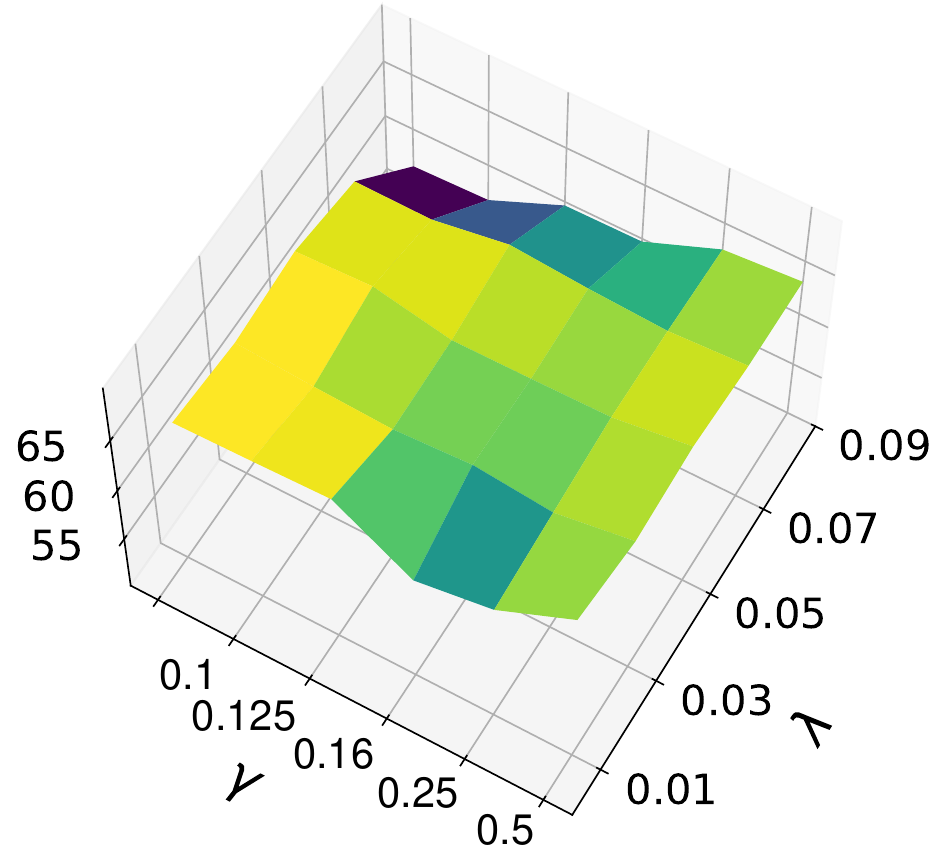}
    \caption{ACC (\%)}
    
  \end{subfigure}
  \vspace{-1em}
  \caption{Hyperparameter analysis under DPGBA on Arxiv.}
  \label{fig:hyper_dpgba}
  \vspace{-1em}
\end{figure}

\subsection{Hyperparameter Analysis}

We further study the effect of the two key hyperparameters: $\lambda_{\text{div}}$ in Eq.~(\ref{eq:logic_loss}), which controls the strength of the MI-based logic diversity loss, and $\gamma$ in Eq.~(\ref{eq:router_obj}), which controls the degree of training router with soft labels for robustness-aware routing.
To explore their effects, we vary the values of $\lambda$ as $\{0.01, 0.03, 0.05, 0.07, 0.09\}$; and $\gamma$ as $\{0.1, 0.125, 0.16, 0.25, 0.5\}$.
Figure~\ref{fig:hyper_dpgba} reports results on Arxiv under the DPGBA attack. We observe that once $\lambda_{\text{div}}>0.03$, {\method} consistently achieves robust performance with ASR$<$20\%. Under some parameters, {\method} can reach near-zero ASR.
Meanwhile, increasing $\lambda$ has little impact on the clean accuracy, implying the logic diversity can enhance the robustness without sacrificing the performance. On the other hand, increasing $\gamma$ can bring some benefit in robustness, with the observed decreased ASR. Its impact on the clean accuracy is also small.


%% file: 2_Related.tex
\section{Related Work}

\noindent \textbf{Graph Adversarial Attacks and Defense}.
Extensive studies have revealed that GNNs are vulnerable to adversarial attacks via edge manipulation \cite{zugner_adversarial_2019, geisler2021robustness}, node feature perturbation \cite{sun2020adversarial, zugner2018adversarial}, and node injection \cite{zou2021tdgia, chenunderstanding2022}. For example, PRBCD~\cite{geisler2021robustness} can fool GNNs to give false predictions on target nodes by adding or deleting an unnoticeable amount of edges of the training graph. TDGIA~\cite{zou2021tdgia} manages to significantly reduce the global node classification performance of GNNs by injecting a small amount of labeled fake nodes. Emerging studies~\cite{dai2023unnoticeable, zhang2024rethinking, zhang2024robustness} also highlighted the significance of graph backdoors.
GTA \cite{xi2021graph} adopts a backdoor trigger generator to generate adaptive sample-specific triggers to improve the attack success rate.
UGBA \cite{dai2023unnoticeable} 
controls the model’s prediction
by injecting specifically designed unnoticeable triggers to selected nodes, achieving high attack success rate under a limited attack budget. DPGBA \cite{zhang2024rethinking} further introduces an outlier detector and uses adversarial learning to generate in-distribution triggers.
In response to edge manipulation and node injection attacks, researchers have proposed a range of defense methods for GNNs including adversarial training~\cite{xu2019topology, zhang2022chasing}, graph structure denoising~\cite{zhu2019robust, jin2020graph}, and robust aggregation~\cite{chenunderstanding, geisler2021robustness, jin2021node}. Although the defense against backdoor attacks has also been recently noticed \cite{zhang2024rethinking, zhang2024robustness}, the defense against backdoor attacks is still rather limited. There lacks a unified framework to defend against various types of adversarial attacks\looseness=-1.

\noindent \textbf{Mixture of Experts}. 
The Sparse Mixture of Experts (MoE) \cite{jacobs1991adaptive, shazeer2017outrageously} architecture determines a group of experts for each input sample, therefore greatly increasing the model's capacity without losses in efficiency. 
Despite the success of MoE in scaling up pre-trained models~\cite{shazeer2017outrageously, fedus2022switch, dai2024deepseekmoe},  existing studies have also investigated the potential of MoE architecture beyond efficiency. In image domain, \citet{puigcerver2022adversarial} observed improved robustness against adversarial attacks by adopting MoE models. 
In the graph domain, \citet{wu2024graphmetro} designed GraphMetro specifically tailored to address complex distribution shifts in graphs. \citet{liu2023fair} diversified the representations of graph experts to learn fair representations. \citet{yuan2024mitigating} further combines the differential privacy and MoE to increase the robustness against node injection attacks. Different from previous papers, our work firstly discovers the diversity and flexibility of MoE to develop a dynamic framework against various graph adversarial attacks.
Please refer to Appendix \ref{app:related_work} for the detailed version of related works.

\vspace{-0.6em}

%% file: 5_Conclusion.tex
\section{Conclusion and Future Work}

In this paper, we identify and excavate two key properties of the MoE architecture: the {diversity} in capturing representations and the {flexibility} in routing samples to selected experts. Building on the natural characteristics of MoE, we introduce two novel designs (i.e., logic diversity loss and robustness-aware router) to propose {\method} against various state-of-the-art adversarial attacks including backdoor, node injection and edge manipulation. 
Moreover, the robustness benefit can
be achieved without significantly increasing inference-time costs, bringing a robust and scalable solution for advancing GNNs' safety in unpredictable real-world applications. In future work, we will explore the broader potential of MoE beyond efficiency, aiming to build GNNs that are both more trustworthy and powerful as well.

%% file: 6_Appendix.tex
\section{Auxiliary loss of {\method}}
\label{app:auxiliary_loss}

\noindent
Following \cite{shazeer2017outrageously}, we define the \emph{importance} of each expert as the batch-wise sum of its gate values:
\begin{equation}
I_k = \sum_{v \in \mathcal{V}} f_\phi(v)_k,
\quad 
\bar{I} = \frac{1}{N}\sum_{k=1}^{N} I_k,
\end{equation}
where $\mathcal{V}$ denotes the batch of nodes and $N$ is the total number of experts. 
The importance loss $\mathcal{L}_{\text{importance}}$ is computed as the squared coefficient of variation (CV) over all experts:
\begin{equation}
\mathcal{L}_{\text{importance}} = 
\frac{\frac{1}{N}\sum_{k=1}^{N}(I_k - \bar{I})^2}{\bar{I}^2}.
\end{equation}

\noindent
To further ensure a balanced assignment of samples across experts, we adopt a \emph{load-balancing loss}~\cite{shazeer2017outrageously}:
\begin{equation}
\mathcal{L}_{\text{load}} = 
\frac{N \cdot \sum_{k=1}^{N} I_k L_k}
{\big(\sum_{k=1}^{N} I_k\big)\big(\sum_{k=1}^{N} L_k\big)},
\end{equation}
where $L_k$ denotes the expected number of samples routed to expert $k$. 
These two auxiliary losses encourage all experts to be both equally important and equally utilized, preventing expert imbalance and training collapse.

\begin{algorithm}[ht] 
\caption{Algorithm of {\method}.} 
\label{alg:rgmoe} 
\begin{algorithmic}[1]
\REQUIRE Poisoned Graph $\mathcal{G}=(\mathcal{V}, \mathcal{E}, \mathbf{X})$, $\mathcal{Y}_L$, $N$, $K$. $\lambda$, $\omega$.
\ENSURE MoE model $f_{\theta}$, robust router $f_\phi$ and MI estimator $T_{\omega}$.
\STATE Randomly initialize $f_{\theta}$, $f_\phi$ and $T_{\omega}$;
\WHILE{not converged yet}
    \STATE Update $f_{\theta}$, $f_\phi$ and $T_{\omega}$ by descent on $\nabla_{\theta, \phi, \omega} (\mathcal{L}_{\text{c}} + \lambda \mathcal{L_{\text{logic}}} )$ based on Eq. (\ref{eq:phase1});
\ENDWHILE
\STATE Identify $\hat{\mathcal{V}}_P$ via expert disagreement based on Eq.(\ref{eq:identify_vp});
\STATE Obtain smooth labels $\tilde{y}$ for $\hat{\mathcal{V}}_P$ based on Eq.(\ref{eq:smooth_y}) ;
\WHILE{not converged yet}
    \STATE Update $f_\phi$ by descent on $\nabla_{\phi}\mathcal{L}_{\text{{router}}}$ based on Eq. (\ref{eq:router_obj});
\ENDWHILE
\RETURN $f_\theta$ and $f_\phi$;
\label{algorithm}
\end{algorithmic}
\end{algorithm}

\section{Training Algorithm of {\method}}

The training algorithm of {\method} is proposed in Algorithm~\ref{alg:rgmoe}. The whole training process is composed of two phases. Specifically, we first train the diverse GNN experts with the initial router from line 1 to line 5 based on Eq.~(\ref{eq:phase1}). Then we fine-tune to get the robustness-aware router by identifying $\hat{\mathcal{V}}_P$ and constructing the smooth labels. The router is optimized from line 7 to line 9 using Eq.~(\ref{eq:router_obj}).

\section{Detailed Related Work on Defense against Graph Adversarial attacks}
\label{app:related_work}

As GNNs are vulnerable to adversarial attacks, various robust graph neural networks against
adversarial attacks have been proposed~\cite{zhang2019graph}. GCN-Jaccard \cite{wu2019adversarial} defends against adversarial attacks by eliminating the edges connecting nodes with low Jaccard similarity of node features. 
RobustGCN \cite{zhu2019robust} is to model Gaussian distributions as hidden
layers to absorb the effects of adversarial attacks in the variances. Inspired by the fact that adversarial attacks will lead to high-rank adjacency matrix, ProGNN \cite{jin2020graph} proposes to learn a low-rank clean adjacency matrix where the adversarial edges are likely to be removed. GNNGuard \cite{zhang2020gnnguard} computes the attention scores based on the cosine similarity of node representations from last layer. With the similarity-based attention, the adversarial edges
are likely to be assigned with small weights since they generally link dissimilar nodes. In MedianGNN \cite{chenunderstanding}, a median aggregation mechanism is designed to improve the
robustness of GNNs. RSGNN \cite{dai2022towards} adopts
the node attributes and supervision from the noisy edges to denoise and dense graph, ensuring the robustness with noisy edges and limited labels. 
\citet{dai2021nrgnn} introduces to learn a robust GNN on sparse and noisy labels. \citet{dai2023unified} proposes a unified information bottleneck framework for robustness against adversarial attack and membership inference.

To address the susceptibility of GNNs to the emerging backdoor attacks, basic defense methods have been recently proposed to remove the trigger from the poisoned dataset. Prune \cite{dai2023unnoticeable} reomve edges that connect nodes with low similarity.  OD \cite{zhang2024rethinking} which involves training a graph auto-encoder and filtering out nodes with high reconstruction loss. RIGBD \cite{zhang2024robustness} identifies the large prediction variance of poisoned nodes under edge dropping and designs a convolution operation to distinguish poisoned nodes from clean nodes. They also propose a novel training strategy to train a backdoor-robust GNN model.
However, the majority of robust GNNs focus on the general manipulation attacks/node injection attacks. The defense on graph backdoor is rather limited let alone a unified framework to defend them simultaneously.

\begin{table}[ht]
\centering
\caption{Dataset Statistics}
\begin{tabular}{lcccc}
\toprule
Datasets & Nodes & Edges & Feature & Classes \\
\midrule
Cora     & 2,708  & 5,429   & 1,443    & 7 \\
Pubmed   & 19,717 & 44,338  & 500      & 3 \\
Flickr   & 89,250 & 899,756 & 500      & 7 \\
OGB-arxiv & 169,343 & 1,166,243 & 128    & 40 \\
\bottomrule
\end{tabular}
\label{tab:datasets}
\end{table}
\begin{table*}[t]
\centering
\small
\caption{Comparison with baselines in defending against manipulation attacks on Cora dataset.}
\vspace{-1em}
\begin{tabularx}{0.98\linewidth}{ll *{8}{>{\centering\arraybackslash}X}}
\toprule
\textbf{Ptb} & \textbf{Dataset} & \textbf{GCN} & \textbf{RGCN} & \textbf{GNNGuard} & \textbf{SoftMedian} & \textbf{SimPGCN} & \textbf{DP-MoE} & \textbf{Ours} \\

\midrule

 3 \% & Cora   & 72.09 $\pm$ 0.00 & 73.81 $\pm$ 0.09 & $76.43 \pm 0.83$ & 77.70 $\pm$ 0.98 & 77.20 $\pm$ 1.00 & $73.51 \pm 0.41$   & 76.65 $\pm$ 0.44 \\
\midrule
5 \% & Cora   & 65.50 $\pm$ 0.57 & 68.15 $\pm$ 0.17 & $74.05 \pm 0.40$ & 74.55 $\pm$ 0.61 & 70.92 $\pm$ 0.09 & $66.41 \pm 2.58$   & 70.06 $\pm$ 0.54 \\
 \midrule
10 \%  & Cora   & 55.95 $\pm$ 0.23 & 58.60 $\pm$ 0.52 & $66.07 \pm 0.38$ & 66.91 $\pm$ 0.80 & 63.46 $\pm$ 0.09 & $56.58 \pm 0.69$  & 64.88 $\pm$ 1.06 \\
\bottomrule
\end{tabularx}
\label{tab:prbcd_cora}
\end{table*}

\begin{table*}[t]
\centering
\small
\caption{Comparison with baselines in defending against node injection attacks on Cora dataset.}
\vspace{-1em}
\begin{tabularx}{0.98\linewidth}{ll *{8}{>{\centering\arraybackslash}X}}
\toprule
& \textbf{Dataset} & \textbf{GCN (clean)} & \textbf{GCN} & \textbf{RGCN} & \textbf{GNNGuard} & \textbf{SoftMedian} & \textbf{SimPGCN} & \textbf{DP-MoE} & \textbf{Ours} \\
\midrule
 & Cora   & 85.34 $\pm$ 0.46 & 69.75 $\pm$ 0.61 & 78.74 $\pm$ 0.54 & $80.45\pm 0.09$ & 77.88 $\pm$ 0.23 & 77.02 $\pm$ 0.46 & $78.80 \pm 0.53$ & 80.59 $\pm$ 0.69 \\
\bottomrule
\end{tabularx}
\label{tab:tdgia_cora}
\end{table*}
\begin{table*}[t]
\centering
\small
\caption{Additional ablation across datasets under GTA.}
\vspace{-1em}
\begin{tabularx}{0.98\linewidth}{ll *{8}{>{\centering\arraybackslash}X}}
\toprule
\multirow{2}{*}{Method} &
\multicolumn{2}{c}{Cora} &
\multicolumn{2}{c}{Pubmed} &
\multicolumn{2}{c}{Flickr} &
\multicolumn{2}{c}{OGB} \\
\cmidrule(r){2-3} \cmidrule(r){4-5} \cmidrule(r){6-7} \cmidrule(r){8-9}
& ASR & CA & ASR & CA & ASR & CA & ASR & CA \\
\midrule
wo Both 
& $79.56 \pm 3.16$ & $89.88 \pm 0.76$ & $72.50 \pm 1.29$ & $88.31 \pm 0.45$ & $100.00 \pm 0.00$ & $49.39 \pm 0.31$ & $79.64 \pm 0.88$ & $61.17 \pm 0.31$
 \\
wo Re-routing 
& $72.89 \pm 4.07$ & $84.56 \pm 2.97$ & $72.05 \pm 1.40$ & $88.13 \pm 0.65$ & $70.58 \pm 11.35$ & $37.67 \pm 1.90$ & $78.69 \pm 1.83$ & $60.53 \pm 0.19$
 \\
wo Div
& $75.41 \pm 2.36$ & $91.48 \pm 0.30$ & $2.06 \pm 1.91$ & $89.58 \pm 0.62$ & $100.00 \pm 0.00$ & $49.21 \pm 1.53$ & $18.35 \pm 18.72$ & $62.46 \pm 1.54$
\\
Ours
& $0.74 \pm 0.76$ & $85.68 \pm 3.50$ & $1.52 \pm 0.94$ & $87.47 \pm 0.22$ & $0.02 \pm 0.01$ & $42.11 \pm 2.11$ & $3.24 \pm 1.42$ & $57.50 \pm 1.36$
 \\
\bottomrule
\end{tabularx}
\label{tab:ablation-gta}
\end{table*}

\begin{table*}[t]
\centering
\small
\caption{Additional ablation across datasets under DPGBA.}
\vspace{-1em}
\begin{tabularx}{0.98\linewidth}{ll *{8}{>{\centering\arraybackslash}X}}
\toprule
\multirow{2}{*}{Method} &
\multicolumn{2}{c}{Cora} &
\multicolumn{2}{c}{Pubmed} &
\multicolumn{2}{c}{Flickr} &
\multicolumn{2}{c}{OGB} \\
\cmidrule(r){2-3} \cmidrule(r){4-5} \cmidrule(r){6-7} \cmidrule(r){8-9}
& ASR & CA & ASR & CA & ASR & CA & ASR & CA \\
\midrule
wo Both 
& $77.78 \pm 3.57$ & $91.48 \pm 0.30$ & $82.20 \pm 0.18$ & $82.36 \pm 0.59$ & $77.34 \pm 6.62$ & $44.57 \pm 1.42$ & $97.56 \pm 0.56$ & $61.29 \pm 1.58$ \\

wo Re-routing 
& $61.19 \pm 2.72$ & $86.17 \pm 2.81$ & $25.25 \pm 2.47$ & $79.55 \pm 1.44$ & $76.09 \pm 5.82$ & $44.87 \pm 1.63$ & $2.00 \pm 0.27$ & $61.34 \pm 0.93$ \\

wo Div
& $75.41 \pm 4.54$ & $91.23 \pm 0.63$ & $82.00 \pm 2.71$ & $81.11 \pm 0.63$ & $6.31 \pm 8.92$ & $45.60 \pm 2.81$ & $90.87 \pm 7.25$ & $60.79 \pm 1.56$ \\

Ours
& $0.30 \pm 0.42$ & $88.27 \pm 4.06$ & $0.62 \pm 0.33$ & $82.61 \pm 1.70$ & $0.01 \pm 0.01$ & $45.14 \pm 2.55$ & $0.02 \pm 0.02$ & $62.11 \pm 0.48$ \\

\bottomrule
\end{tabularx}
\label{tab:ablation-dpgba}
\end{table*}

\section{Dataset Statistics}
\label{app:datasets}

\noindent \textbf{Cora and Pubmed} are citation networks where nodes denote papers, and edges depict citation relationships.
In Cora, each node is described using a binary word vector,
indicating the presence or absence of a corresponding word from a predefined dictionary.
In contrast, PubMed employs a TF/IDF weighted word vector for each node. Nodes are categorized based on their respective research areas for both datasets.

\noindent \textbf{Flickr} represents images uploaded to Flickr as nodes in a graph. An edge is created between two nodes if the corresponding images share certain attributes, such as location, gallery, or user comments. Each image is described by a 500-dimensional bag-of-words feature vector derived from the NUS-wide dataset. Labels are based on 81 tags assigned to the images, which were manually grouped into 7 distinct classes. Each image is categorized into one of these classes.

\noindent \textbf{OGB-Arxiv} is a citation network consisting of all Computer Science papers from arXiv cataloged in the Microsoft Academic Graph. Each paper is represented by a node, with a 128-dimensional feature vector derived by averaging the skip-gram word embeddings from the paper's title and abstract. The nodes are classified according to their respective research areas.
The statistics details of these
datasets are summarized in Tab.~\ref{tab:datasets}.

\section{Implementation Details}
\label{app:implement_details}

\noindent \textbf{Train, validation, and test splits.} In this paper, we conduct semi-supervised node classification experiments. For backdoor attacks, 20\% of nodes are selected as training nodes on Pubmed, Flickr, and OGB-Arxiv, and 70\% on Cora due to its smaller size. 10\% are used as the validation set, while 20\% are masked as test nodes, split equally for ASR (10\%) and ACC (10\%) evaluation. For manipulation and node injection attacks, 20\% of nodes are randomly selected as training nodes across all datasets, with 10\% for validation and 20\% masked as test nodes.

\noindent \textbf{Attack budgets for manipulation and node injection.} For PRBCD, we evaluate different perturbation rates (3\%, 5\%, and 10\%) as shown in Tables \ref{tab:prbcd} and \ref{tab:prbcd_cora}. For TDGIA, we inject 40 malicious nodes with 3 edges each on Cora, 5000 malicious nodes with 10 edges each on Flickr, and 1000 malicious nodes with 50 edges each on OGB-Arxiv.

\noindent \textbf{Parameter configuration.} Following previous studies~\cite{dai2023unnoticeable}, we use a 2-layer GCN as each MoE expert as well as the gating model. The hidden dimension is set to 32 for all datasets. The learning rate is set to 0.01 with 5e-4 weight decay and we use an Adam optimizer to train 200 epochs for all baseline methods. All experiments are conducted on 8 A6000 GPUs.

\label{app:robustness_of_graphmoe}
\begin{figure}[ht]
  \vspace{-1em}
  \centering
  \includegraphics[width=0.5\linewidth]{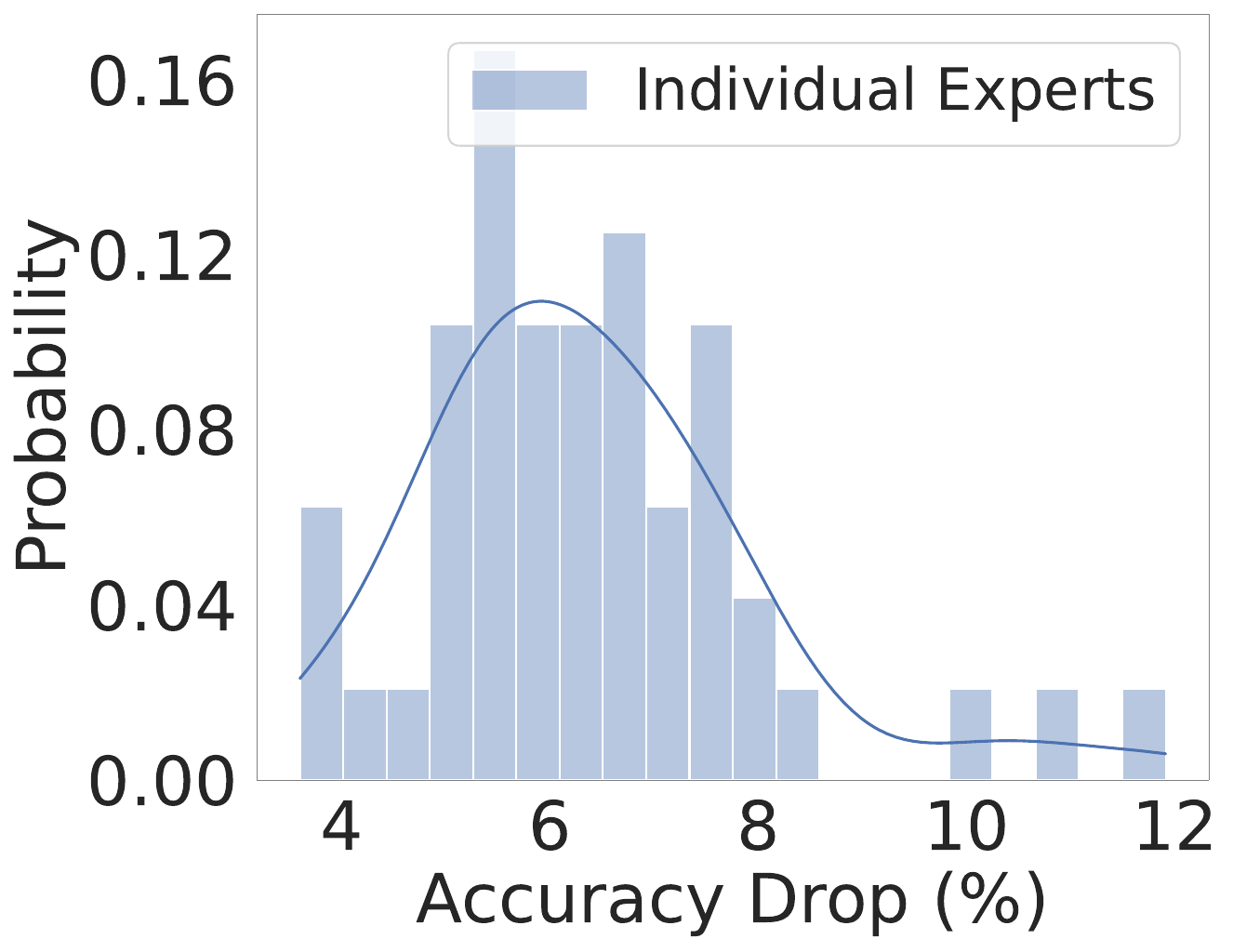}
  \caption{Performance distribution of individual experts in Graph MoE under node injection attack.}
  \label{fig:aps}
\end{figure}

\label{app:distribution_compare}
\begin{figure}[ht]
  \centering
  \begin{subfigure}[t]{.22\textwidth}
    \centering
    \includegraphics[width=\linewidth]{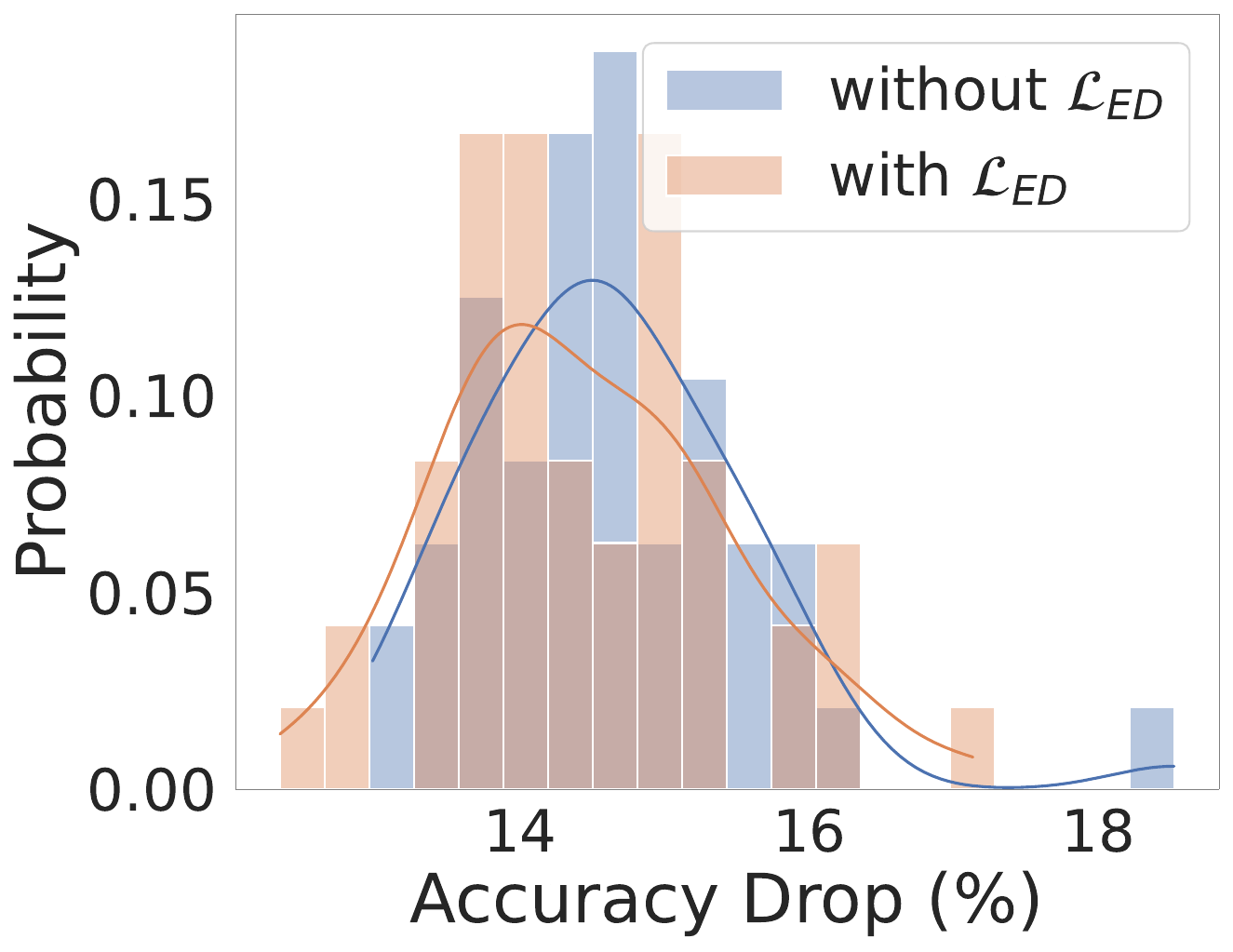}
    \caption{PRBCD}
  \end{subfigure}\hfill
  \begin{subfigure}[t]{.22\textwidth}
    \centering
    \includegraphics[width=\linewidth]{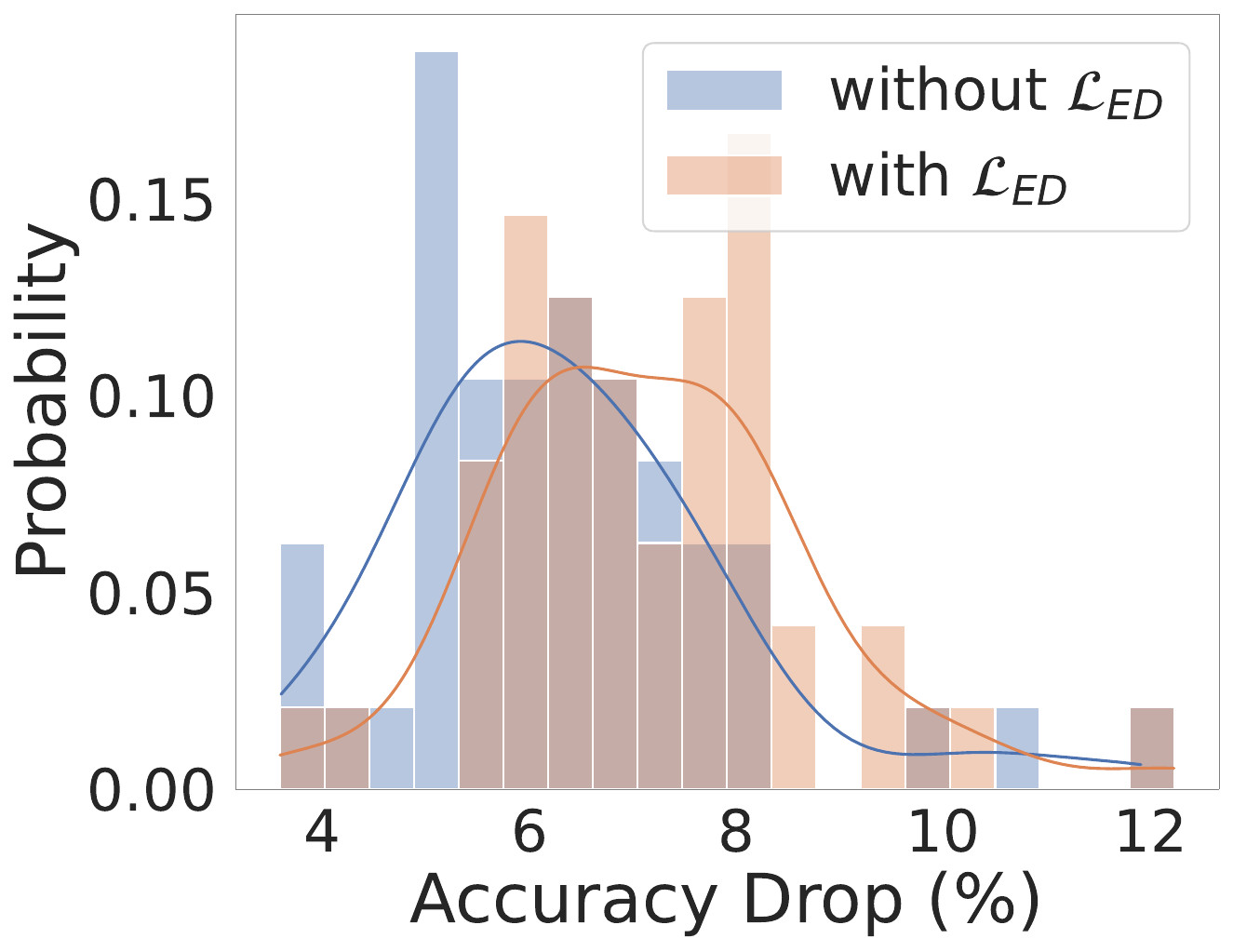}
\caption{TDGIA}
  \end{subfigure}\hfill
  \vspace{-1em}
  \caption{Impacts of existing diversity loss on the proportion of robust individual experts under PRBCD and TDGIA.}
  \label{fig:diversity_distribution_app}
  \vspace{-1 em}
\end{figure}

\vspace{-0.6em}

\section{Additional robustness analysis under diversity regularization}
\label{app:diversity_distribution}
In this section, we give the formulation of the representation diversity loss $\mathcal{L}_{ED}$ proposed in~\cite{liu2023fair} as well as its impact on experts' robustness distribution under PRBCD and TDGIA attacks. Given a node $v$ and its $L$-hop neighbors, this loss enriches the diversity of learned node embeddings by maximizing the agreement between its neighborhood nodes  while pushing away the irrelevant nodes that are
far away. The procedure is formulated as follows:
\begin{equation}
    \mathcal{L}_{ED} = -\text{log}\frac{\sum_{j\in \mathcal{N}(v)}\text{exp}(\text{sim}(z_v, z_j))}{\sum_{k\notin \mathcal{N}(v)}\text{exp}(\text{sim}(z_v, z_k))},
\end{equation}
where $z$ represents the learned node embedding. The function sim() calculates the similarity between two vectors, i.e.:
\begin{equation}
    \text{sim}(z_i, z_j) = \frac{z_i^{\mathrm{T}} z_j}{||z_i||_2||z_j||_2}.
\end{equation}
The impact of $\mathcal{L}_{ED}$ to the robustness of individual experts is illustrated in Fig.~\ref{fig:diversity_distribution_app}. In addition, Fig.~\ref{fig:aps} illustrates the results under TDGIA without the this diversity regularization.

\section{Additional Experimental Results}

\noindent \textbf{Additional Results of Edge Manipulation and Node Injection Attacks}.
\label{app:results_adv}
In this subsection, we report the additional experimental results under PRBCD and TDGIA on the small Cora dataset. The results are reported in Tab.\ref{tab:prbcd_cora} and Tab.~\ref{tab:tdgia_cora}, respectively.

\begin{figure}[ht]
  \centering
  \begin{subfigure}[t]{.48\columnwidth}
    \centering
    \includegraphics[width=\linewidth]{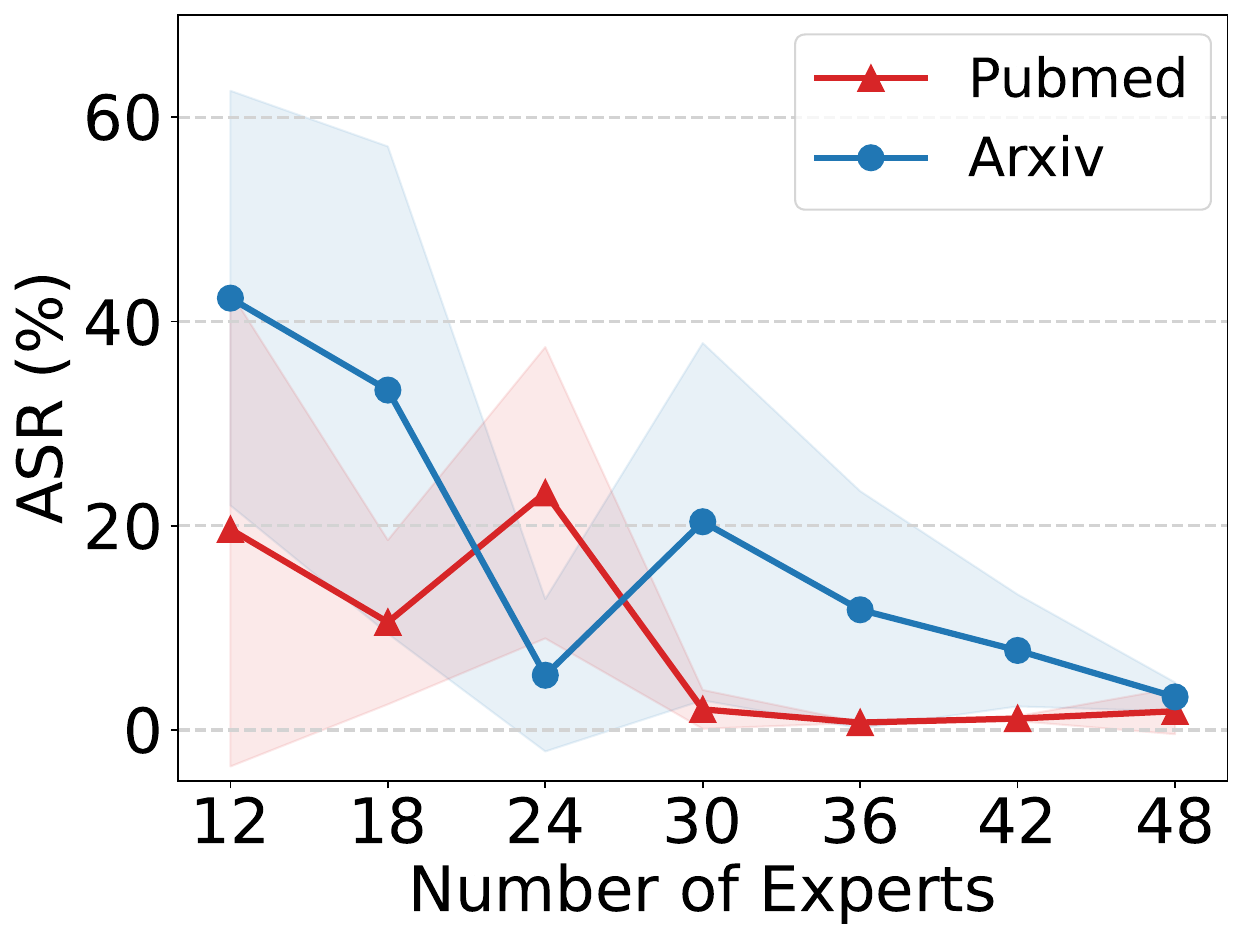}
  \end{subfigure}
  \hfill
  \begin{subfigure}[t]{.48\columnwidth}
    \centering
    \includegraphics[width=\linewidth]{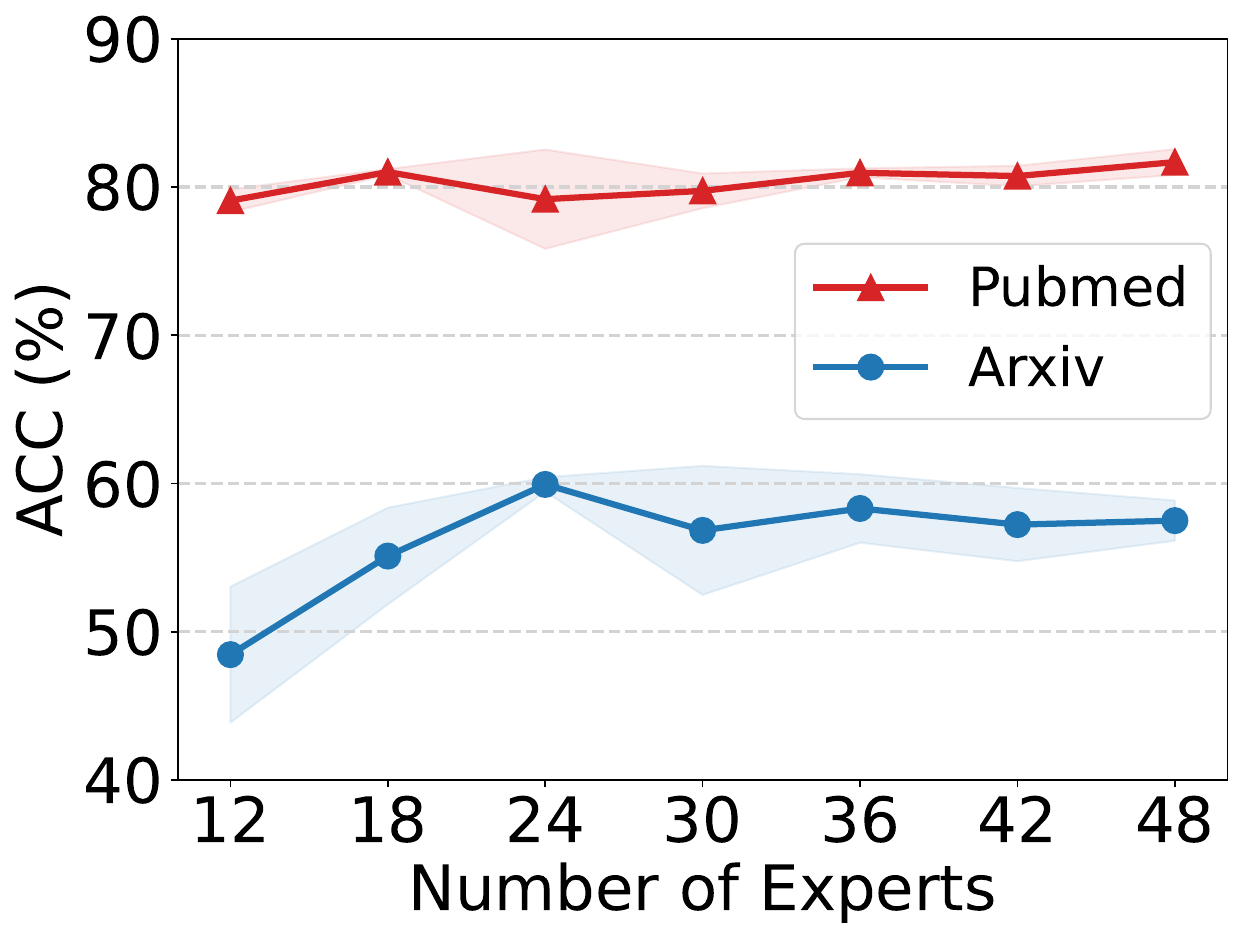}
  \end{subfigure}
  \vspace{-2mm}
  \caption{Impact of model capacity under GTA attack.}
  \label{fig:capacity_gta}
\end{figure}

\begin{figure}[ht]
  \small
  \centering
  \begin{subfigure}[t]{.48\columnwidth}
    \centering
    \includegraphics[width=\linewidth]{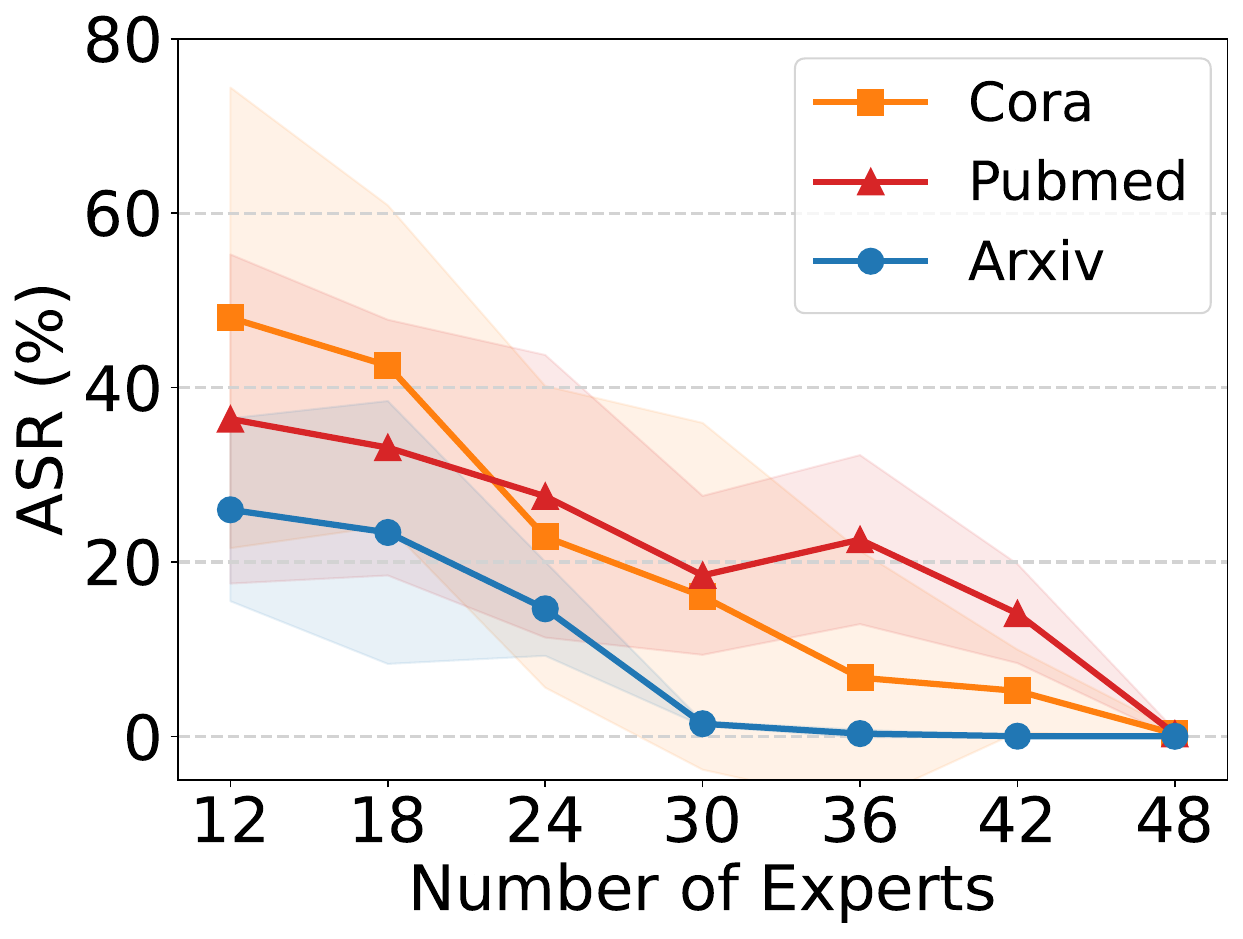}
  \end{subfigure}
  \hfill
  \begin{subfigure}[t]{.48\columnwidth}
    \centering
    \includegraphics[width=\linewidth]{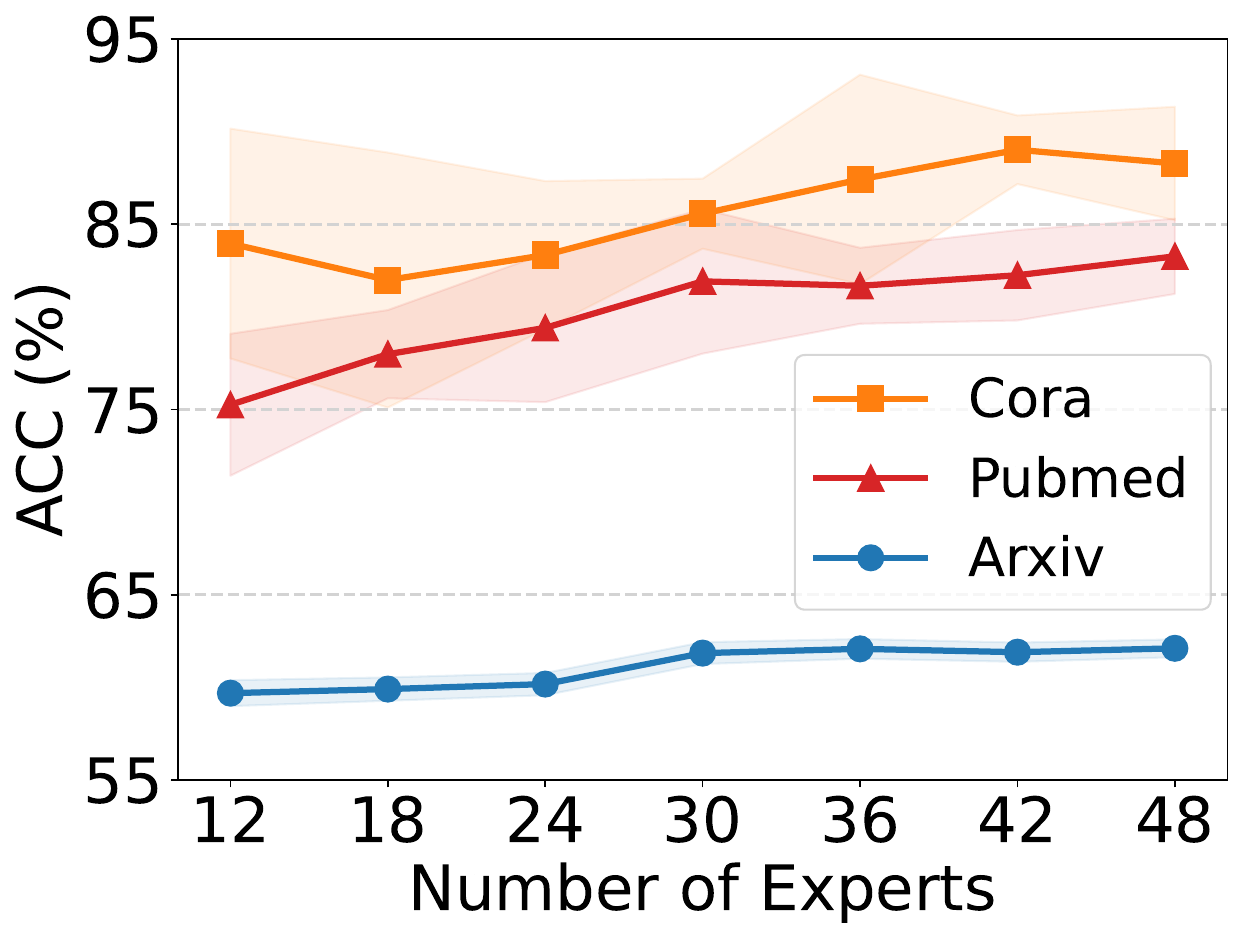}
  \end{subfigure}
  \vspace{-2mm}
  \caption{Impact of model capacity under DPGBA attack.}
  \label{fig:capacity_DPGBA}
\end{figure}

\noindent \textbf{Additional Results of Impact of Model Capacity}.
\label{app:capacity_results}
In this subsection, we report the additional results of impact of model capacity under GTA and DPGBA attack in Fig.~\ref{fig:capacity_gta} and Fug.~\ref{fig:capacity_DPGBA}, respectively.

\begin{figure}[h]
  \centering
  \begin{subfigure}[t]{.48\columnwidth}
    \centering
    \includegraphics[width=\linewidth]{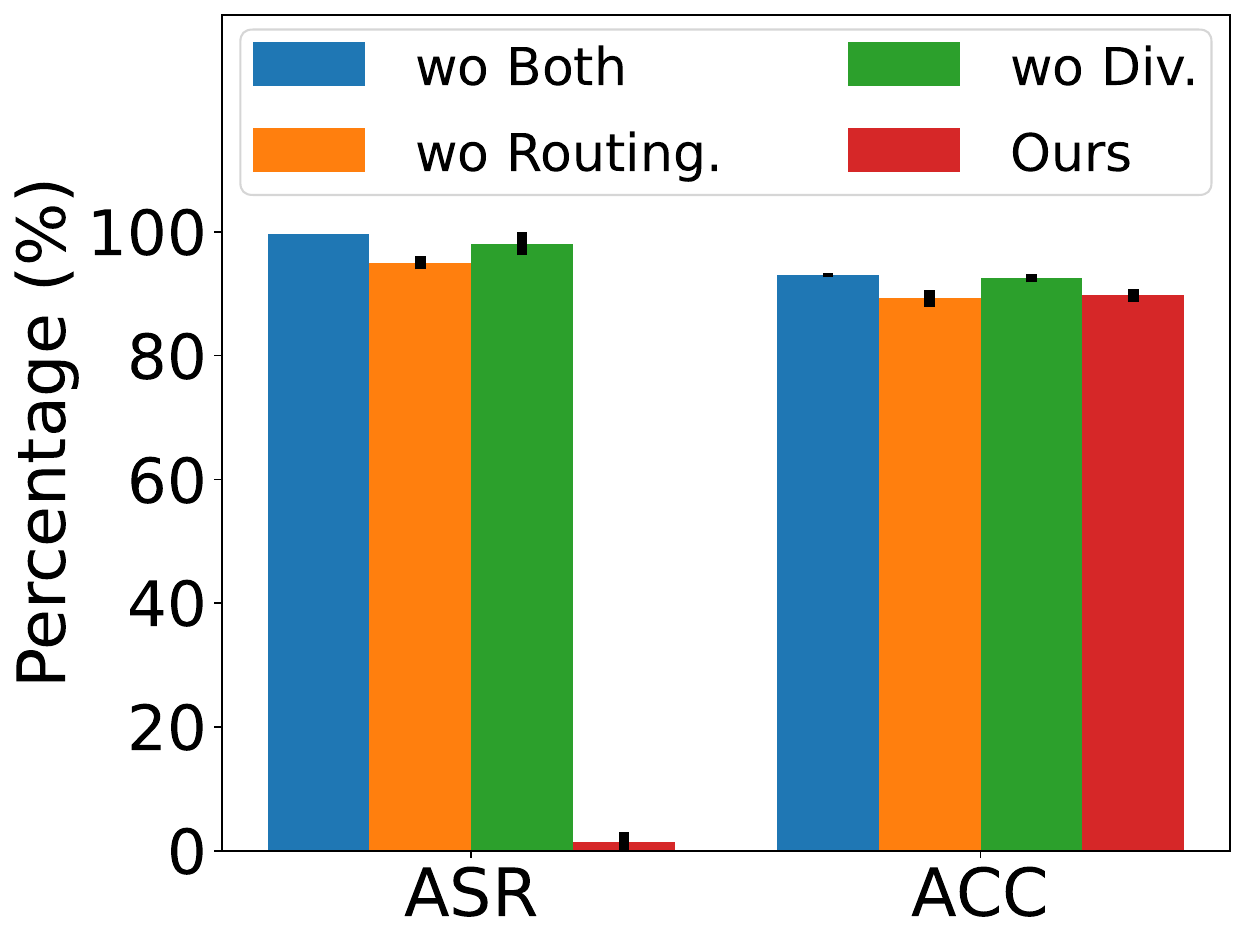}
\caption{Cora}
  \end{subfigure}
  \hfill
  \begin{subfigure}[t]{.48\columnwidth}
    \centering
    \includegraphics[width=\linewidth]{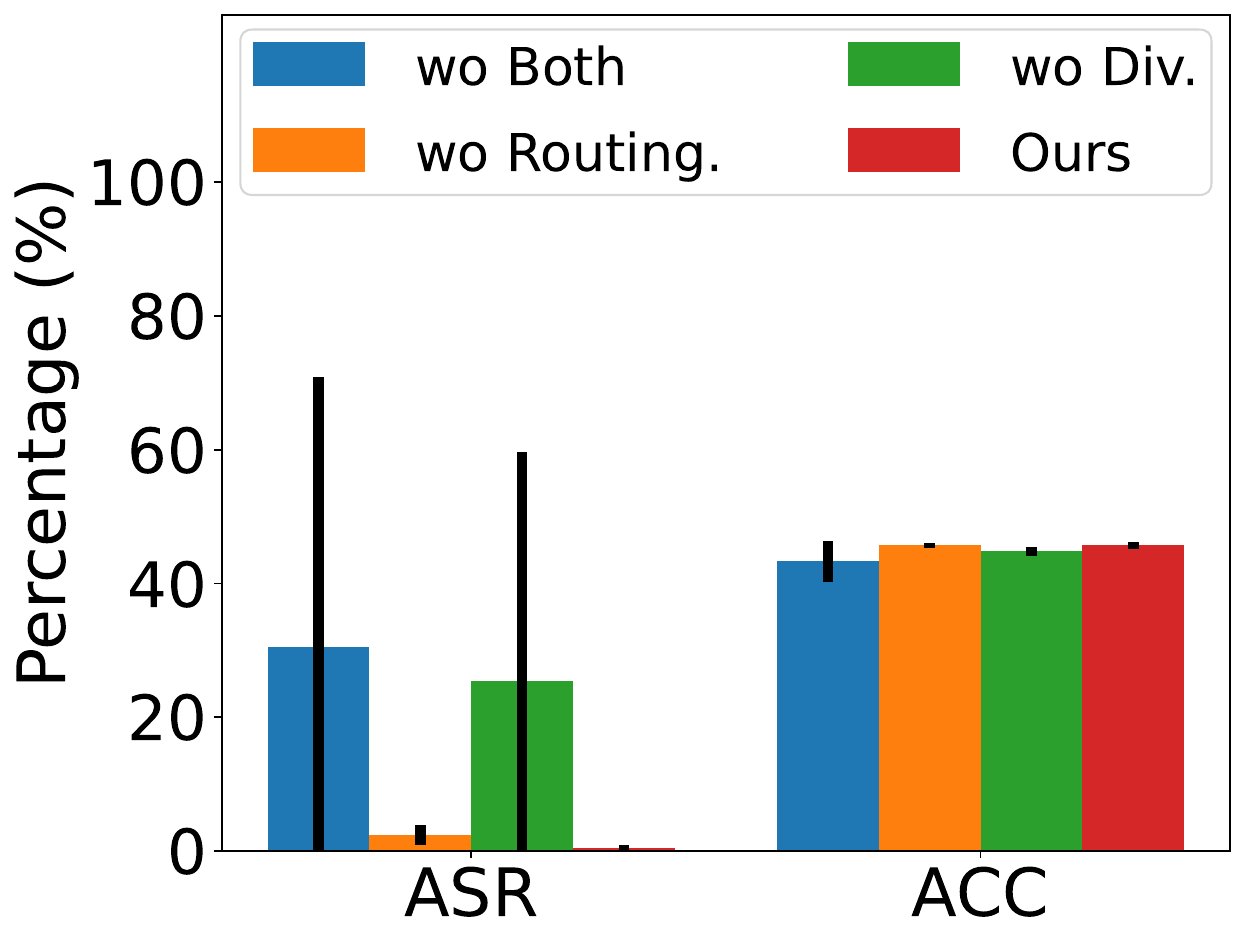}
    \caption{Flickr}
  \end{subfigure}
  \vspace{-2mm}
  \caption{Additional ablation on UGBA across datasets.}
  \label{fig:additional_abl_ugba}
\end{figure}

\noindent \textbf{Additional Results of Ablation Study}.
\label{app:ablation_results}
In this subsection, we report the additional ablation study results on Cora and Flickr under UGBA in Fig.~\ref{fig:additional_abl_ugba}. The ablation results under other attacks across the datasets are reported in Tab.~\ref{tab:ablation-gta} and Tab.~\ref{tab:ablation-dpgba}.

\newpage 

